\documentclass[twocolumn]{article}
\usepackage[utf8]{inputenc}
\usepackage{amssymb}
\usepackage{latexsym}
\usepackage{textcomp}
\usepackage{array}
\usepackage{amsmath}
\usepackage{tabularx}
\usepackage{multirow}
\usepackage{scrextend}
\usepackage{tikz}
\usepackage{collcell}
\usepackage{etoolbox}
\usepackage{url}
\usepackage{authblk}
\usepackage{xcolor}
\usepackage{hyperref}
\usepackage{textcomp}  
\usepackage{array}
\usepackage{tabularx}
\usepackage{scrextend}
\usepackage{tikz}
\usepackage{collcell}
\usepackage{etoolbox}

\newcommand{\PUCP}[0]{PUCP}%
\newcommand{\HCMUS}[0]{HCMUS}%
\newcommand{\baseline}[0]{Baseline}
\newcommand{\Bhref}[3][blue]{\href{#2}{\color{#1}{#3}}}%

\newcommand{\PreserveBackslash}[1]{\let\temp=\\#1\let\\=\temp}
\newcolumntype{C}[1]{>{\PreserveBackslash\centering}p{#1}}

\newtoggle{inTableHeader}
\toggletrue{inTableHeader}
\newcommand*{\StartTableHeader}{\global\toggletrue{inTableHeader}}%
\newcommand*{\EndTableHeader}{\global\togglefalse{inTableHeader}}%

\let\OldTabular\tabular%
\let\OldEndTabular\endtabular%
\renewenvironment{tabular}{\StartTableHeader\OldTabular}{\OldEndTabular\StartTableHeader}%

\newcommand*{\MinNumber}{0.4}%
\newcommand*{\MidNumber}{0.7} %
\newcommand*{\MaxNumber}{0.8}%

\newcommand{\ApplyGradient}[1]{%
  \iftoggle{inTableHeader}{#1}{
    \ifdim #1 pt > \MidNumber pt
        \pgfmathsetmacro{\PercentColor}{max(min(100.0*(#1 - \MidNumber)/(\MaxNumber-\MidNumber),100.0),0.00)} %
        \hspace{-0.33em}\colorbox{green!\PercentColor!yellow}{#1}
    \else
        \pgfmathsetmacro{\PercentColor}{max(min(100.0*(\MidNumber - #1)/(\MidNumber-\MinNumber),100.0),0.00)} %
        \hspace{-0.33em}\colorbox{red!\PercentColor!yellow}{#1}
    \fi
  }}

\newcolumntype{R}{>{\collectcell\ApplyGradient}c<{\endcollectcell}}
\begin{document}

\title{SHREC 2022: pothole and crack detection in the road pavement using images and RGB-D data}

\author[1]{Elia Moscoso Thompson}
\author[1]{Andrea Ranieri}
\author[1]{Silvia Biasotti}
\author[2]{Miguel Chicchon}
\author[3]{Ivan Sipiran}

\author[4,5]{Minh-Khoi Pham}
\author[4,5]{Thang-Long Nguyen-Ho}
\author[4,5,6]{Hai-Dang Nguyen}
\author[4,5,6]{Minh-Triet Tran}

\affil[1]{\small Istituto di Matematica Applicata e Tecnologie Informatiche `E. Magenes' - CNR}
\affil[2]{Pontificia Universidad Catolica Del Per$\acute{u}$}
\affil[3]{Department of Computer Science, University of Chile}

\affil[4]{University of Science, Ho Chi Minh City, Vietnam}
\affil[5]{Viet Nam National University, Ho Chi Minh City, Vietnam}
\affil[6]{John von Neumann Institute, Ho Chi Minh City, Vietnam}
\normalsize

\twocolumn[
  \begin{@twocolumnfalse}
    \maketitle
    \begin{abstract}
        This paper describes the methods submitted for evaluation to the SHREC 2022 track on pothole and crack detection in the road pavement. A total of 7 different runs for the semantic segmentation of the road surface are compared, 6 from the participants plus a baseline method. All methods exploit Deep Learning techniques and their performance is tested using the same environment (i.e.: a single Jupyter notebook). A training set, composed of 3836 semantic segmentation image/mask pairs and 797 RGB-D video clips collected with the latest depth cameras was made available to the participants. The methods are then evaluated on the 496 image/masks pairs in the validation set, on the 504 pairs in the test set and finally on 8 video clips. The analysis of the results is based on quantitative metrics for image segmentation and qualitative analysis of the video clips. The participation and the results show that the scenario is of great interest and that the use of RGB-D data is still challenging in this context.
        
        \bigskip
        \bigskip
        \bigskip
    \end{abstract}
  \end{@twocolumnfalse}
]
\maketitle

\section{Introduction}
\label{sec:introduction}
Road infrastructure is one of the most important foundations of modern society. The interconnection between cities and towns is important both for the transport of people and goods. The road network continues to be the solution that best combines cost and efficiency to reach locations that would otherwise not be reached by the rail network. However, its main constructive component, the asphalt, tends to deteriorate considerably with time, use and atmospheric events (e.g. rain, snow, frost, etc.). To repair this kind of damage, constant and complete monitoring of the road infrastructure is necessary but, due to the high costs, it is often neglected or delayed over time to the detriment of the quality of the road surface. Furthermore, the monitoring of road sections alone, verifying when it is necessary to intervene and what type of intervention is required, is expensive and impractical. Indeed, the scheduling of inspections and maintenance is entrusted to specialized personnel who require specific training and operate expensive and bulky machinery~\cite{DU2021}. Overall, data from US authorities indicates that currently the expenses for both vehicle damages (related to road mismanagement) and road maintenance are in the order of billions USD/year~\cite{tripreport2018}. This is a significant bottleneck for those in charge of road maintenance that can be avoided with technologies aimed at improving and automating these tasks, reducing human effort and costs. 

It is, therefore, no surprise that the interest in the topic of road pavement analysis has recently grown and many high-quality works~\cite{DU2021} have been produced.
In this contest, we focus our attention on two kinds of road damage: \emph{cracks} and \emph{potholes}. In the contest of this paper, we consider the following concepts: 
\begin{itemize}
    \item Cracks: one or multiple fractures in the road surface. The length of cracks tends to always exceed their width by orders of magnitude.
    \item Potholes: a portion of asphalt that is missing or crumbled to the point of having a significant displacement in the surface (i.e.: the inside of a pothole is lower than the rest of the road surface) and/or the terrain under the road surface is clearly visible. 
\end{itemize}
In our context, the main difference between a crack and a pothole is width rather than depth.

In this SHREC track, we compare methods that automate crack and pothole detection by enabling timely monitoring of large areas of road pavement through the use of Deep Learning (DL) techniques.
The goal is to recognize and segment potholes and cracks in images and videos using a training set of images enriched by RGB-D video clips. For completeness, it is worth mentioning that other kinds of data can be used when working with road-related tasks. For example, Ground Penetrating Radar (GPR) data is generated using electromagnetic waves to scout what is on and below the road surface (e.g.:~\cite{TONG2017}) but this data source requires very expensive equipment and specialized personnel to operate.

This paper is organized as follows. In Section~\ref{sec:related_works} we summarize the state of the art regarding road damage datasets, while in Section~\ref{sec:contest_setup} we describe the datasets, the task in detail and the numeric evaluation measures used in this SHREC contest. In Section~\ref{sec:methods} we summarize the methods evaluated in this contest, while their performances are described and discussed in Section~\ref{sec:results_discussion}. Finally, conclusion and final remarks are in Section~\ref{sec:conclusions}.

%
%
%
\section{Related datasets}
\label{sec:related_works}
The problem of road damage detection using image-based techniques has gained great importance in the last 15 years with the explosion of Computer Vision and Pattern Recognition methods. This rapid growth has led to the publication of numerous surveys comparing different methods, such as~\cite{Kim2014ReviewAA,Zakeri2017,ali2022bibliometric}. The proposed methods vary in terms of the type of data analyzed and the approach. For example, in~\cite{KOCH2011507} the authors propose an image segmentation method based on histograms and thresholds, then, to detect potholes, they further analyze each segment using texture comparison. Another example is~\cite{Yamaguchi2010}, in which the authors proposed a high-speed crack detection method employing percolation-based image processing.

However, due to the nature of our work and the prospect of being able to use cheap acquisition techniques, we focus more on the literature related to DL methods.
Modern DL techniques have begun to require ever-larger datasets, composed of thousands of high-resolution images, definitely much more complex to collect for small research groups.
How data is collected is crucial, especially when large amounts need to be collected and labelled. Luckily, it is at least possible to collect road images with a number of different tools, from specialized cameras to mid-to-low end phone cameras. In some works, like in~\cite{Koch2011}, authors even extended their datasets using simple online resources, like the Google image search engine.

In \cite{Majidifard19} authors summarize the availability of datasets at the time and divide them into two categories: wide view and top-down view. The first class consists of images of a large area of road pavement along with other elements (buildings, sidewalks, etc.). Examples of this kind of datasets are presented in~\cite{Pothole-600_4,crack500_1,Maeda18}.

The second class consists of images that are optimal when it comes to assessing damage to the asphalt, as they offer a more accurate view of the road, but at the cost of not representing the entire damaged area (e.g. a large hole that expands beyond the camera's field of view) or to provide a little context about elements surrounding that specific damage (and thus possibly increasing the risk of confusing e.g. a tar stain with a pothole).
However, the tools required to efficiently sample this kind of images are more sophisticated, thus less available and/or more bulky and expensive. To the best of our knowledge, the first freely available dataset of this kind is~\cite{Eisenbach2017}, which used a specialized vehicle to sample ~2000 images of damaged asphalt. Another dataset, based on data delivered by the Federal Highway Administration, that belongs to this class is~\cite{GOPALAKRISHNAN2017}. Regarding \cite{Majidifard19}, it proposes an object detection dataset consisting of more than 14000 samples created using the Google API street view. However, the image quality is not very high and images show numerous artefacts due to the Google Street View stitching algorithm. In more recent times, in~\cite{Dharneeshkar20} authors travelled across India to capture road damages on asphalted, cemented and dirt roads, acquiring about 1500 images using an iPhone 7 camera. Perhaps one of the most complete datasets for object detection is provided in \cite{Arya20}: it is built on pre-existing datasets and consists of approximately 26000 images, with street samples from multiple countries for further heterogeneity.

In our benchmark, we aim to perform semantic segmentation of road images, i.e. detect and classify road cracks and potholes with pixel accuracy. However, the type of ground truth that corresponds to this task is uncommon, as it is very expensive in terms of human labelling time. In fact, most of the aforementioned datasets are annotated using bounding boxes on the objects of interest. This approach speeds up the labelling phase at the cost of being much less precise in locating the object of interest and in evaluating its real size.
To implement our benchmark we looked for datasets whose ground truth allows semantic segmentation: in Section~\ref{sec:dataset}, we describe those of interest for our purposes.

Finally, it is worth discussing RGB-D data as a middle ground between 3D and 2D data. RGB-D provides an easier way to detect road damage, based on the height displacement of the road surface. It also comes with a relatively low barrier to entry in terms of tools needed: in~\cite{Ouma20}, for example, a Kinect v2.0 camera was used to record portions of the road at up to 30 FPS and 300,000 points per frame, which were later used to generate RGB-D images.
RGB-D technology is, therefore, a very convenient way to collect pre-labelled images which then allow performing a full-fledged "unsupervised learning". Quotation marks are mandatory in this case as RGB-D images tend to be noisy, especially in a scenario such as a road surface monitoring where the required height accuracies are often borderline with those provided by modern consumer depth cameras, often limited by a very short baseline.
%
%
%

\section{Benchmark}
\label{sec:contest_setup}
In the following, we describe the data used in the contest, which consists of both images and video data, and the task given to the participants. Then, we explain how we evaluate the results in quantitative terms and, finally, how we qualitatively evaluate them.

\subsection{Dataset and task proposed}
\label{sec:dataset}
The dataset for this contest is called \emph{Pothole Mix} and it consists of an \emph{image dataset} and an \emph{RGB-D video dataset}. The image dataset is composed of 4340 image pairs (made of RGB images+segmentation masks), collected from 5 high quality public datasets as well as a small set of images manually segmented by the organizers. Each dataset had its own unique labelling in the form of segmentation masks, so to make it possible to train DL models over the entire dataset, we uniformed the masks colors. A sample from each image dataset is shown in Figure~\ref{fig:ds_im_examples}. We represent the cracks in \textcolor{red}{\textbf{red}} and the potholes in \textcolor{blue}{\textbf{blue}}. We detail these datasets (and the criteria behind the split in training, validation and test sets) in the following:
\begin{figure*}[t]
    \centering
    \setlength{\tabcolsep}{0pt}
    \newcommand{\datasetimageswidth}{.165}
    \begin{tabular}{||cccccc||}
        \hline \hline
        \multicolumn{2}{||c}{Crack500} &
        \multicolumn{2}{c}{GAPs384} &
        \multicolumn{2}{c||}{EdmCrack600} \\
        \includegraphics[width=\datasetimageswidth\linewidth]{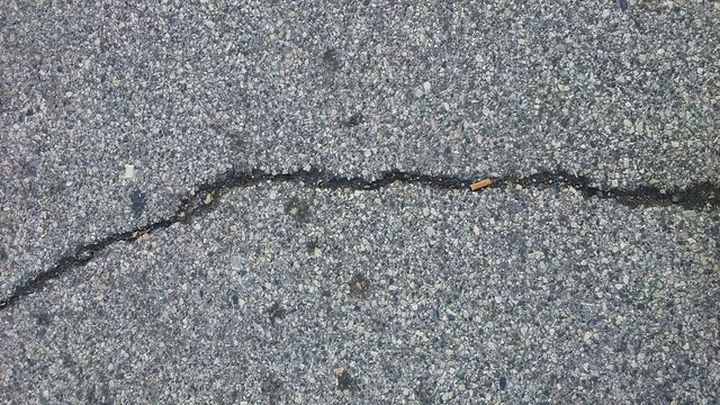} &
        \includegraphics[width=\datasetimageswidth\linewidth]{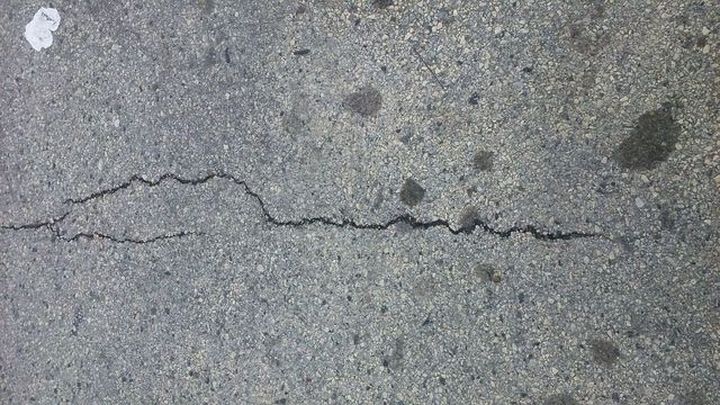} &
        \includegraphics[width=\datasetimageswidth\linewidth]{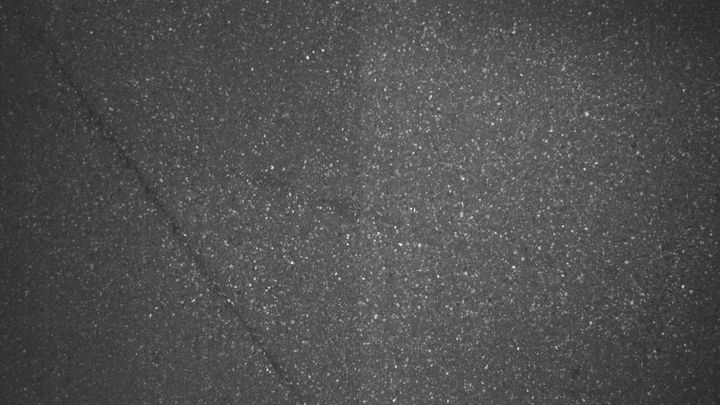} &
        \includegraphics[width=\datasetimageswidth\linewidth]{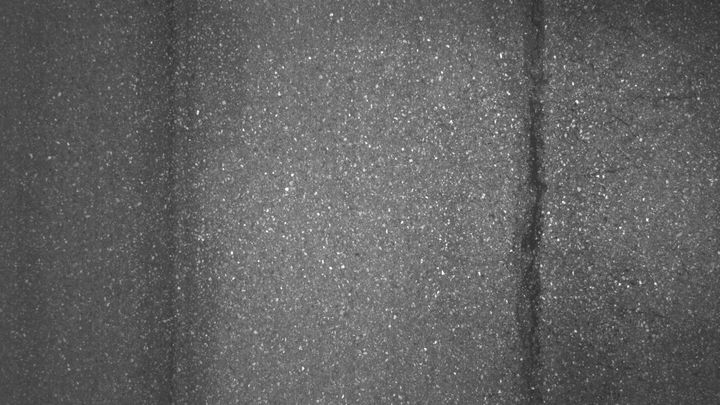} &
        \includegraphics[width=\datasetimageswidth\linewidth]{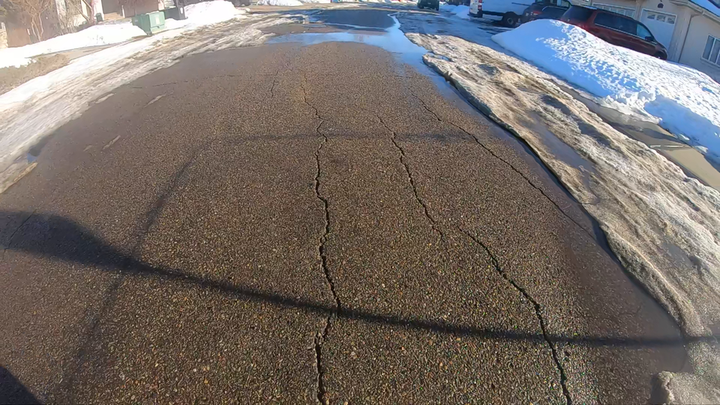} &
        \includegraphics[width=\datasetimageswidth\linewidth]{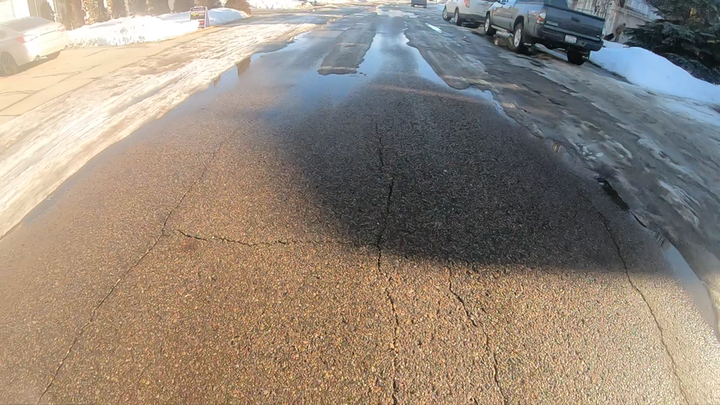} \\
        \includegraphics[width=\datasetimageswidth\linewidth]{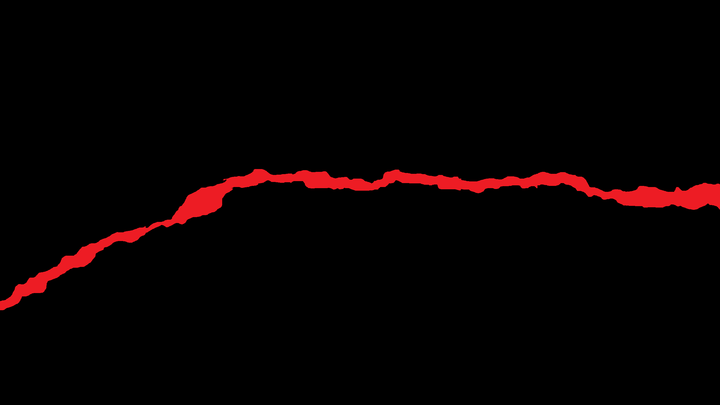} &
        \includegraphics[width=\datasetimageswidth\linewidth]{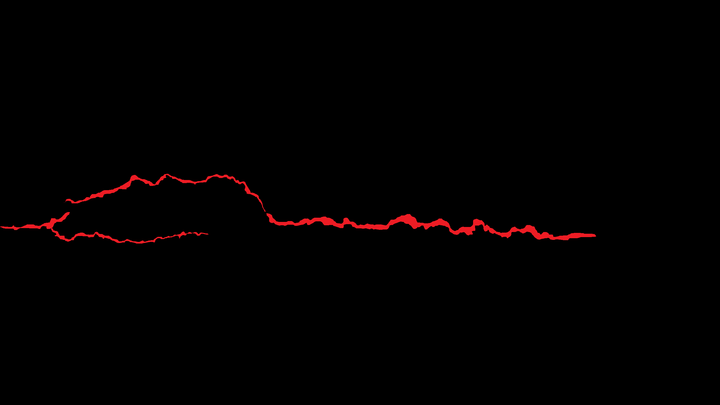} &
        \includegraphics[width=\datasetimageswidth\linewidth]{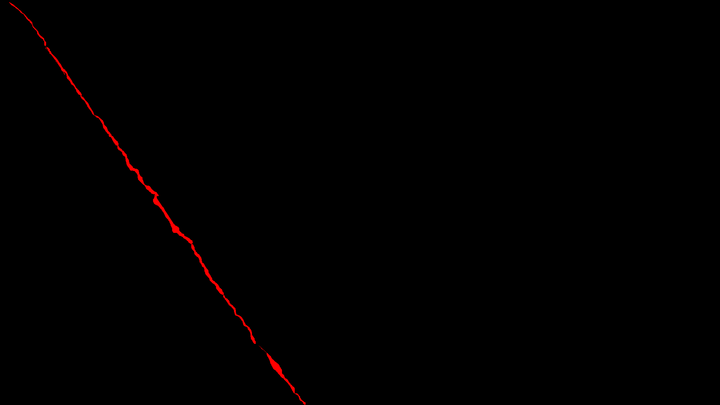} &
        \includegraphics[width=\datasetimageswidth\linewidth]{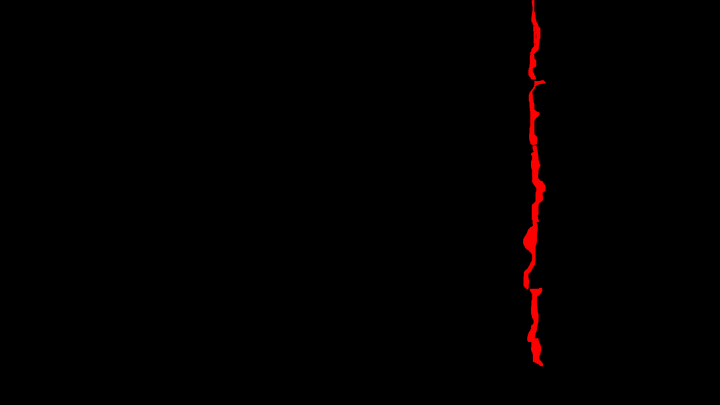} &
        \includegraphics[width=\datasetimageswidth\linewidth]{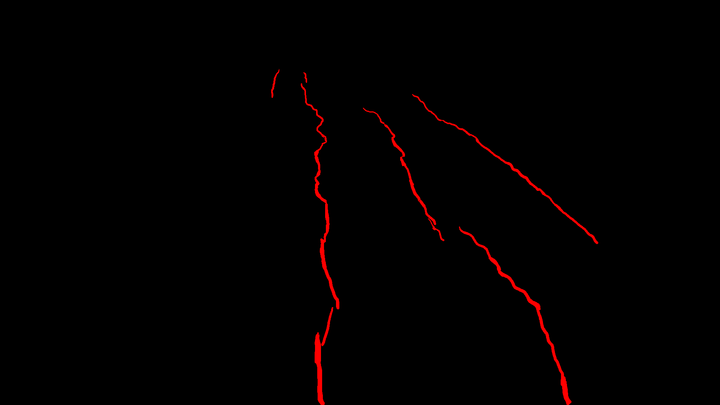} &
        \includegraphics[width=\datasetimageswidth\linewidth]{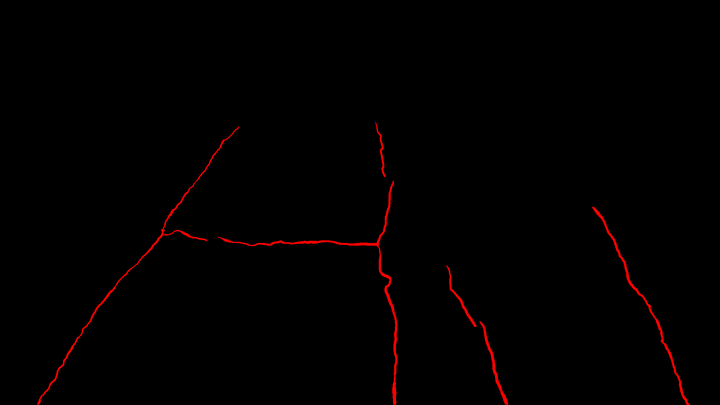} \\ &&&&& \\
        \multicolumn{2}{||c}{Pothole-600} &
        \multicolumn{2}{c}{CPRID} &
        \multicolumn{2}{c||}{Web images} \\
        \includegraphics[trim={0 0 0 5.3cm},clip,width=\datasetimageswidth\linewidth]{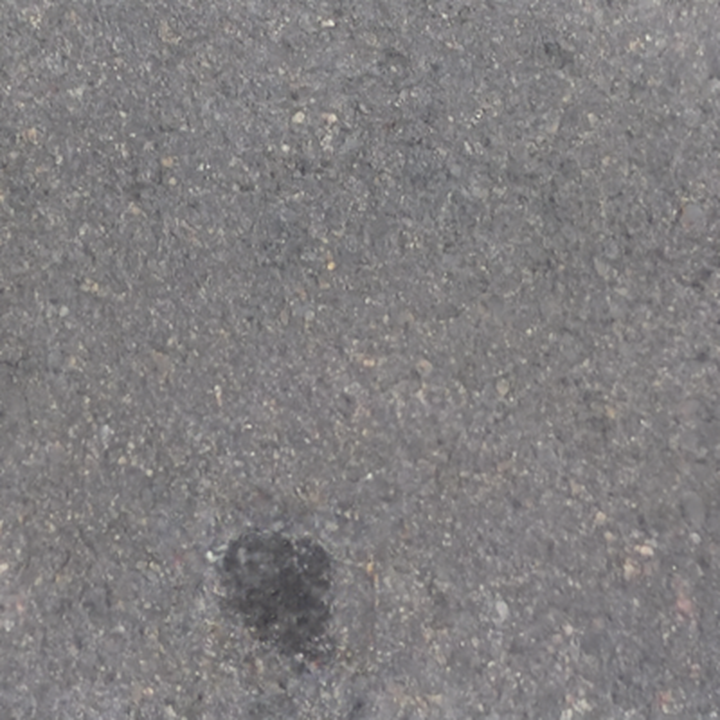} &
        \includegraphics[trim={0 0 0 5.3cm},clip,width=\datasetimageswidth\linewidth]{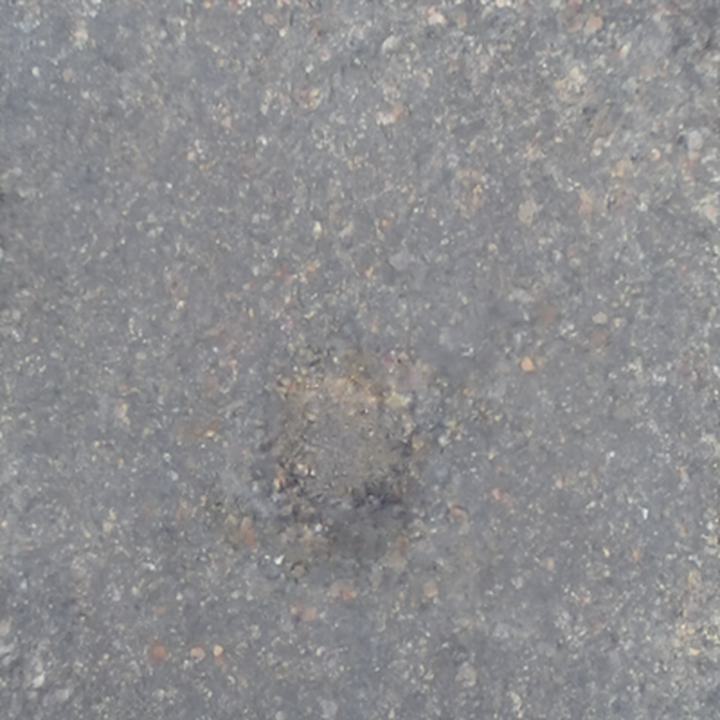} &
        \includegraphics[width=\datasetimageswidth\linewidth]{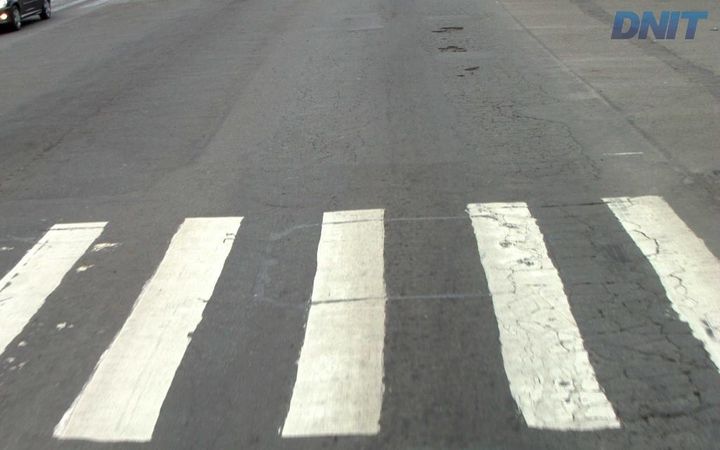} &
        \includegraphics[width=\datasetimageswidth\linewidth]{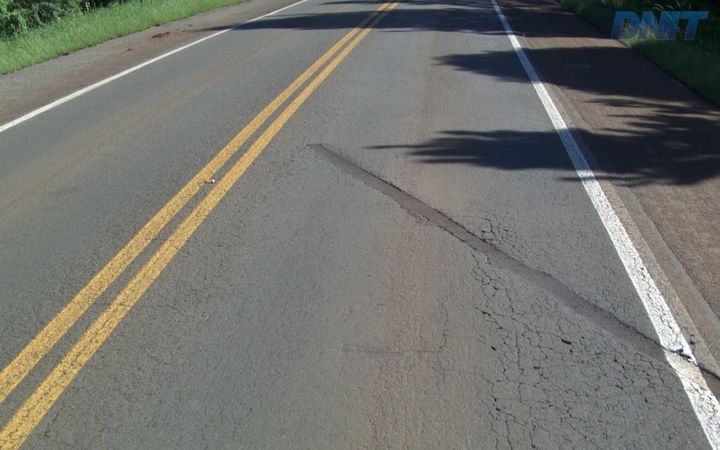} &
        \includegraphics[trim={0 0 0 7.3cm},clip,width=\datasetimageswidth\linewidth]{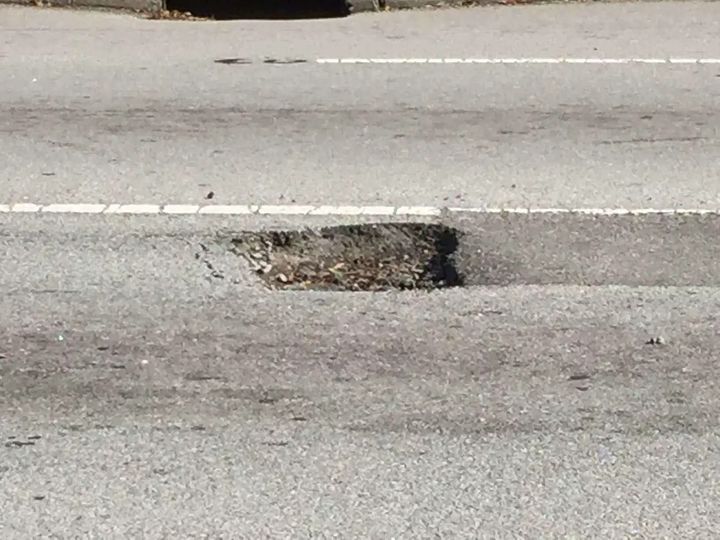} & \includegraphics[trim={0 0 0 1.5cm},clip,width=\datasetimageswidth\linewidth]{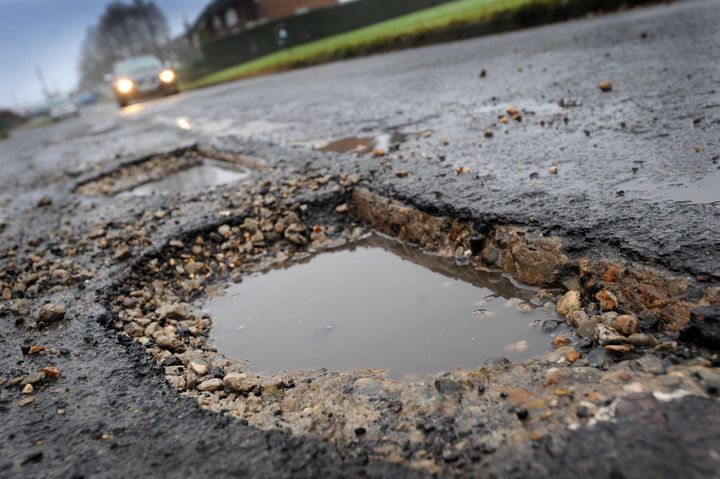} \\
        \includegraphics[trim={0 0 0 5.3cm},clip,width=\datasetimageswidth\linewidth]{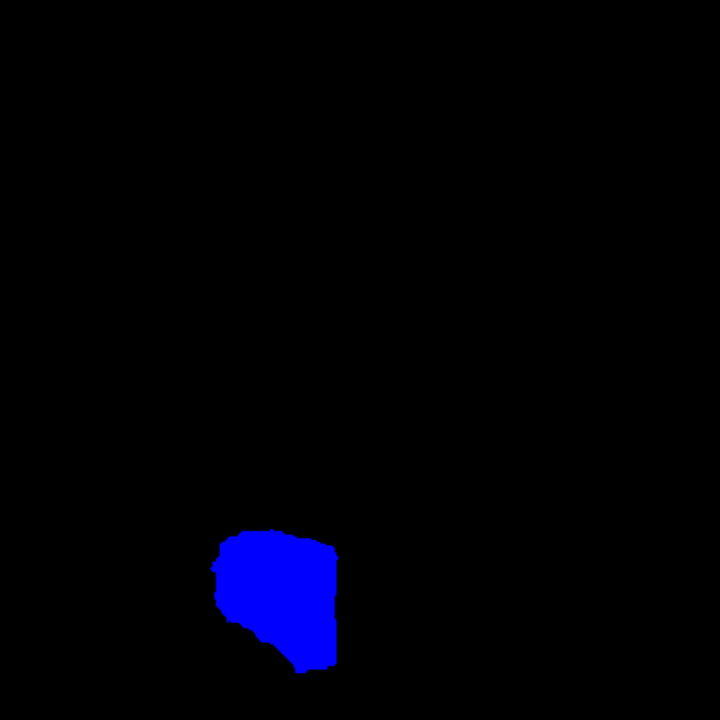} & \includegraphics[trim={0 0 0 5.3cm},clip,width=\datasetimageswidth\linewidth]{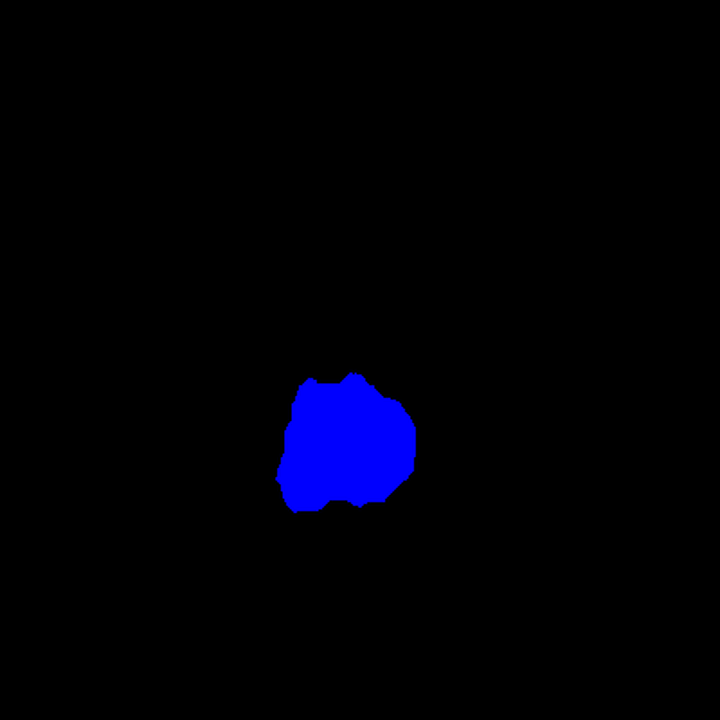} &
        \includegraphics[width=\datasetimageswidth\linewidth]{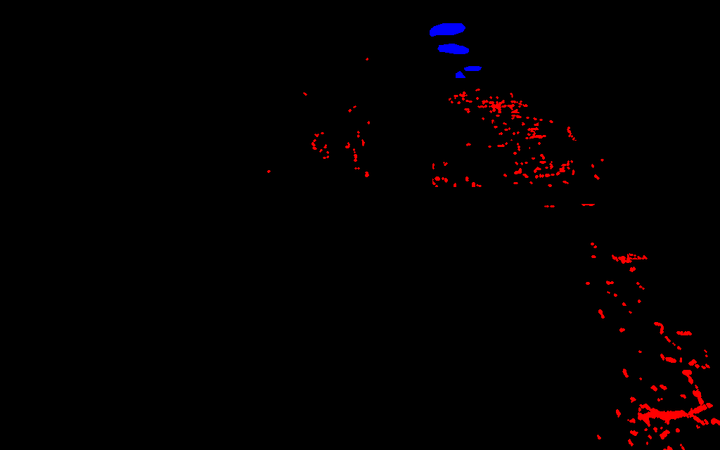} &
        \includegraphics[width=\datasetimageswidth\linewidth]{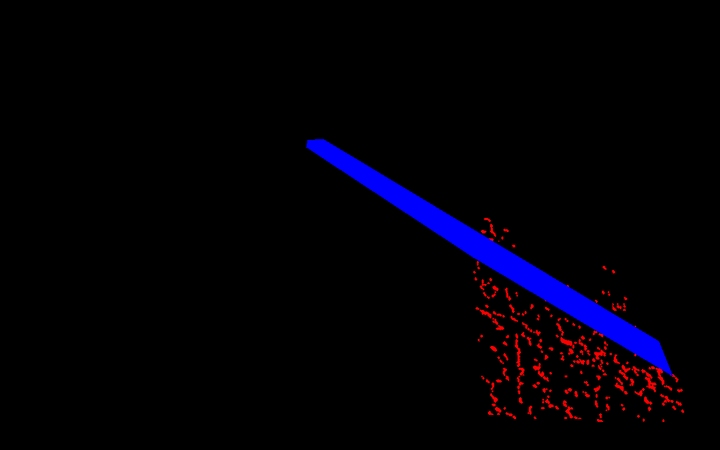} &
        \includegraphics[trim={0 0 0 5.4cm},clip,width=\datasetimageswidth\linewidth]{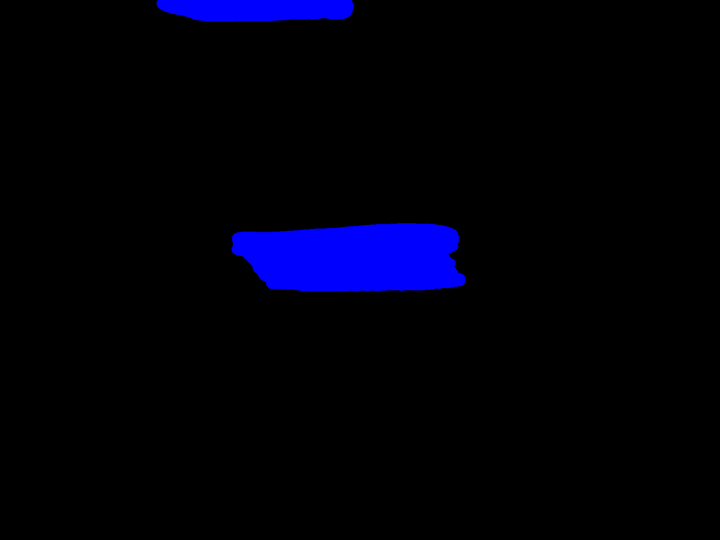} & \includegraphics[trim={0 0 0 6.25cm},clip,width=\datasetimageswidth\linewidth]{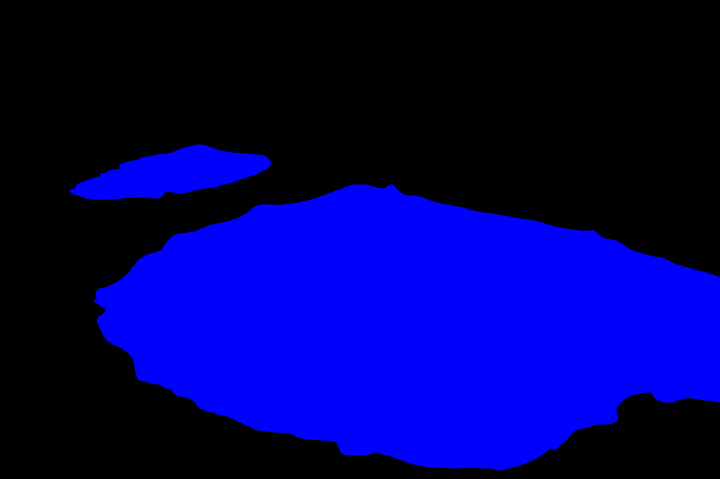} \\ \hline \hline
    \end{tabular}
    \caption{An overview of the images available in the SHREC 2022 benchmark dataset. A couple of samples are drafted from each original dataset and the set of images segmented by us. Below each image, the respective mask is reported. Red indicates cracks, while blue indicates potholes.}
    \label{fig:ds_im_examples}
\end{figure*}
\begin{figure}
    \centering
    \setlength{\tabcolsep}{0pt}
    \begin{tabular}{ccc}
        \includegraphics[width=.33\linewidth]{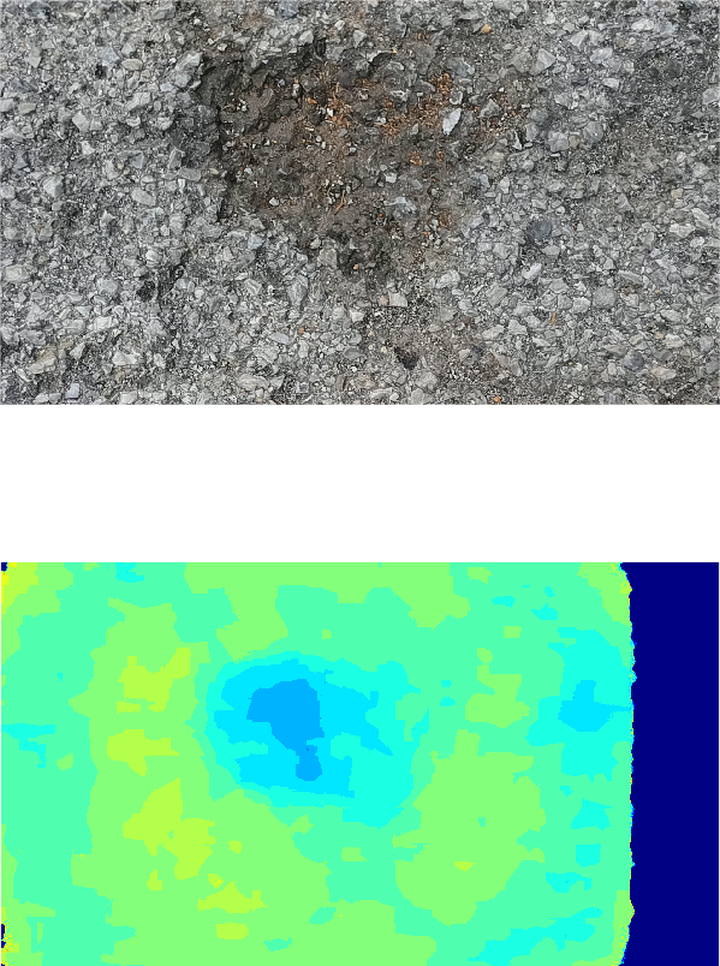} &
        \includegraphics[width=.33\linewidth]{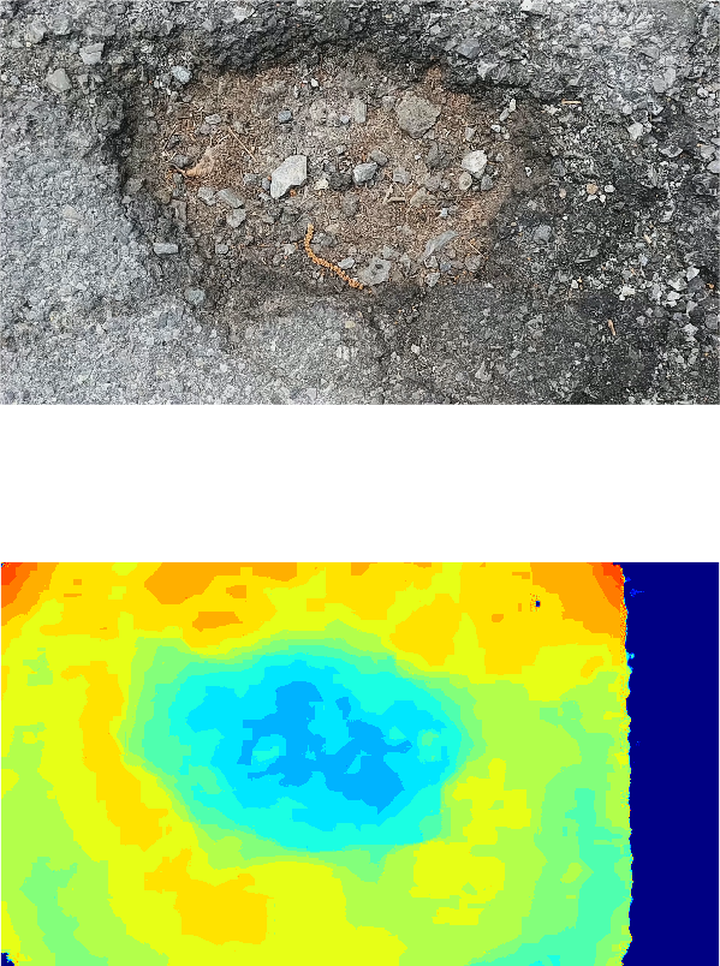} &
        \includegraphics[width=.33\linewidth]{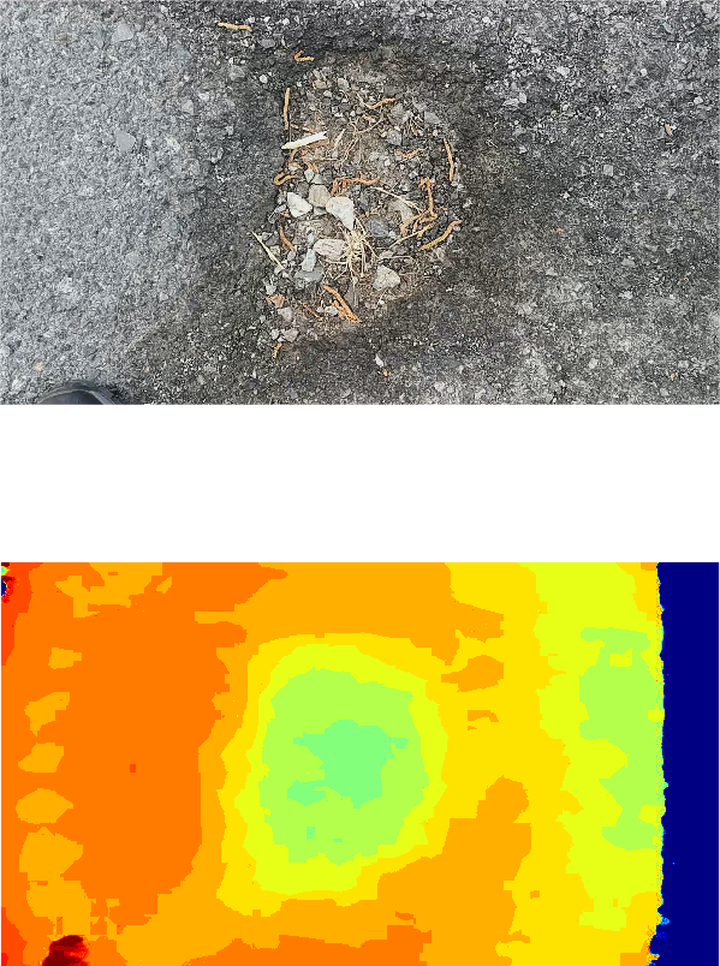}
    \end{tabular}
    \caption{An example of raw frames of three of the clips we captured using the Luxonis OAK-D camera. Below each frame, the respective disparity map is shown in jet colormap (actual disparity videos in the dataset are gray-scale).}
    \label{fig:video_sample}
\end{figure}

\begin{itemize}
    \item Crack500~\cite{crack500_1,crack500_2}: this dataset contains 500 (image/mask) pairs divided in a 250/50/200 split (50/10/40 in percentage). The images have a resolution of $2000\times1500$ px and have been taken from top-down view with cellphones. The images also have the date and time of capture in the file name, were taken from February 22, 2016 to April 15, 2016 and sometimes occur in groups due to spatially close shots. The split is actually random and for this reason all three splits may contain subsets of similar images. This dataset has the peculiarity of incorporating the EXIF metadata coming from the smartphones of origin, so it is necessary to take this into account when loading the images to feed the neural network.
    \item GAPs384: the German Asphalt Pavement distresS~\cite{GAPs384_2} (GAPs384) is a collection of 384 images (out of 1969 total images) with a resolution of $1920\times1080$ px in grayscale with top-down view. The authors in ~\cite{crack500_2} manually selected 384 images from the GAPs dataset which included only cracks, and conducted a pixel-wise annotation on them. The dataset is composed by 353/4/27 image/mask pairs in its training/validation/test sets respectively, giving this dataset a somehow "atypical" split of 92/1/7\%. The images in this dataset are very homogeneous and the training, validation and test sets are derived from sequential images of three distinct road sections that, therefore, have no overlap.
    \item EdmCrack600~\cite{EdmCrack600_1,EdmCrack600_2,EdmCrack600_3}: this dataset was created by capturing images on the streets of Edmonton, Canada and includes 600 pixel-level annotated images of road cracks. Although in the paper the adopted split is random and with a proportion of 420/60/120 pairs (70/10/20 in percentage), the dataset that can be downloaded from the GitHub repository has not been split. For this reason, we decided to randomly split this dataset into 480/60/60 pairs (80/10/10 in percentage) in order to give some more images to the network during the training.
    \item Pothole-600~\cite{Pothole-600_1,Pothole-600_2,Pothole-600_3,Pothole-600_4}: this dataset is made by top-down images collected using a ZED stereo camera that captured stereo road images with a $400\times400$ px resolution. It counts 600 RGB images, the same amount of disparity images and binary segmentation masks. These images have been split by the original authors into training/validation/test sets respectively with a proportion of 240/180/180 (40/30/30 in percentage) and we have kept the same split in this work.
    \item Cracks and Potholes in Road Images Dataset~\cite{CaPiRID} (CPRID): these 2235 images of Brazil highways have been provided by DNIT (National Department of Transport Infrastructure). They were captured in the states of Espirito Santo, Rio Grande do Sul and the Federal District between 2014 and 2017 and were manually selected to be free of vehicles, people or other types of defects in the image. The resolution of the images is $1024x640$ px and the associated ground truth is a segmentation mask to discriminate between cracks and potholes. The dataset is not split so we adopted the split 2000/200/35 images (i.e. 89/9/1 percent) for training/validation/test sets respectively.
    \item Web images: a small set of 20 wide-view high-resolution images of potholes has been retrieved with Google images and annotated with hand-made pixel-perfect semantic segmentation (the split here is 17/2/1).
\end{itemize}

The image dataset as a whole is composed of 4340 image/mask pairs at different resolutions divided into training/validation/test sets with a proportion of 3340/496/504 images equal to 77/11/12 percent.

In addition to images, we provide 797 non-annotated RGB-D video clips (notice that each clip comes with a RGB video and a disparity map video) from which participants can extract additional images to enrich the working dataset. Indeed, we think that the provided disparity maps could help training better models for detecting road damages, since both cracks and potholes correspond to variations in the depth of the road surface, which are visible in the disparity maps. Moreover, even if we provide only short clips, it is possible to extract a large number of images from each of them, given the 15-fps frame rate (see later). We gave no guidelines on how to employ the disparity maps in each clip: we left complete freedom to the participants on how (and if) to use the disparity information provided to improve their methods. These clips are taken with a Luxonis OAK-D camera connected via USB-C to an Android mobile phone using a Unity app. We captured images of the damaged asphalt of extra-urban roads, at varying height (30cm to 1m, according to the size and depth of the pothole). RGB videos are captured in Full HD ($1920\times1080$ px) at 15 FPS (due to mobile phone+app performance limitations). Disparity videos are gray-scale and captured at $640\times400$ px resolution and 15 FPS.
It is worth mentioning that the Luxonis OAK-D camera is able to provide both the disparity image (displacement of each pixel with respect to the two cameras) and depth (real calculation of the 3D position of the points, based on the disparity) of the scene. The camera is also equipped with an Intel Movidius Myriad X processor, capable of running small neural networks to perform inference directly on the device or encode multiple high-resolution, high-frame rate video streams out of the camera. However, while the disparity image is provided at 8 bits and can then be passed to the H.264 or H.265 compression engines, the depth image is provided at 16 bits and thus (at the time of writing this article) it was not possible to create a pipeline with this data flow to be compressed directly on the device.
We therefore opted for the disparity image as the depth videos are captured in an uncompressed format, creating too large amounts of data that we can't comfortably handle with our current setup. The filtering applied directly by the OAK-D camera to each frame of disparity videos consists of a Median Filter with a $7x7$ kernel and another filter based on the confidence returned by the stereo matching algorithm that sets to 0 any pixel under the specified confidence threshold (245 out of 255 in our setup). These clips vary in length, from less than 1 second up to 45 seconds each, and in the type of damage they portray. The disparity map of these videos is noisy and needs denoising before it can become a true segmentation mask, a task that is left to do to the contest participants. Figure~\ref{fig:video_sample} shows a couple of frames from two of these clips. All the data aforementioned is publicly available on Mendeley \Bhref{https://data.mendeley.com/datasets/kfth5g2xk3/1}{at this link}.

The final aim of the task is \emph{to train neural network models capable of performing the semantic segmentation of road surface damage (potholes and cracks)}. 

\subsection{Quantitative measures}
\label{sec:eval_mes}
The quantitative assessment is based on standard metrics on the image dataset. In particular:
\begin{enumerate}
    \item \emph{Weighted Pixel Accuracy (WPA)}: this measure is inspired by~\cite{BrostowFC:PRL2008,BrostowSFC:ECCV08}. In short, it checks how many pixels of a predicted segmentation class are correctly identified as potholes or cracks, without considering the unlabelled pixels in both the ground-truth mask and the predicted one. In our use-case, unlabelled pixels are those depicting undamaged asphalt, painted signposting and other road elements. This metric is designed to give an indication of the "net" pixel accuracy, thus without considering everything that is asphalt (i.e. most of the image).
    \item \emph{Dice Multiclass (DiceMulti)}: it extends the concept of the S{\o}resen-Dice coefficient~\cite{sorenson1948method}, which is two times the area of overlap between a binary mask predicted and its ground-truth divided by the sum of the pixel of both images. In short, Dice multiclass calculates the average of this value for each class, making it a good and widely used evaluation metric for semantic segmentation tasks. See~\cite{DICEMULTI} for more details.
    \item \emph{Intersection over Union (IoU) and mean IoU}: given a binary prediction mask and a binary ground-truth mask, the IoU score is equal to the area (i.e.: number of pixels) of the intersection of the masks over area of the union of the masks. The IoU for a class is the mean across all the samples. Since we are dealing with multiple classes, to obtain the mean of the IoU (mIoU) a confusion matrix has to be built. In this benchmark we use the IoU on potholes alone (pIoU) and cracks alone (cIoU) and the mIoU, ignoring the background also in this metric. 
\end{enumerate}

\subsection{Qualitative evaluation}
\label{sec:qualitative_eval}
Our qualitative evaluation is done on a small set of video clips of road surface, containing cracks, potholes, both or none of them. Our judgment is driven by the visual accuracy of the segmentation, its temporal stability, amount of false positives and false negatives. Given the definitions of cracks and potholes in Section~\ref{sec:introduction}, no particular expertise to assess such a judgement is required. Indeed, while subjective, the organizers were never split in the identification of cracks and potholes. We are confident that, for a qualitative evaluation, common human perception is enough to distinguish between cracks and potholes (or a lack thereof).

\section{Methods}
\label{sec:methods}
Twelve groups registered to this SHREC track but only two teams submitted their results, including the models trained and the code to make it possible to verify them. Each of the two groups sent three submissions for a total of six runs. In the following, we briefly describe how the proposed methods work. We initially introduce a baseline method run by the organizers, then we describe the methods proposed by the participants.


\subsection{Baseline (DeepLabv3+)}
As a baseline, we used the DeepLabv3+~\cite{chen2018encoder} architecture equipped with the a ResNet-101~\cite{resnet101} encoder pre-trained on ImageNet~\cite{deng2009imagenet}, following a similar approach to what was presented in~\cite{fan2020we}. 

Model training took place within a Jupyter Notebook running Python 3.8 and using the popular Fast.ai library now at its second version~\cite{fastai}. Fast.ai adds an additional layer of abstraction above Pytorch~\cite{paszke2019pytorch}, therefore it is very convenient to use to speed up the "standard" and repetitive tasks of training a neural network.

The training exploited the progressive resizing technique~\cite{cellular_super_resolution} ($360\times360$ px $\rightarrow$ $540\times540$ px) in three ways. First, it is exploited as a form of data augmentation. Second, it is used as a methodology to accelerate the convergence of the network on lower resolution images. Finally, the progressive resizing technique allows an early assessment of the quality of the other data augmentations used. In particular, the following data augmentations have been used to postpone overfitting as much as possible: \emph{Blur, CLAHE, GridDistortion, OpticalDistortion, RandomRotate90, ShiftScaleRotate, Transpose, ElasticTransform, HorizontalFlip, HueSaturationValue}. 
In order to maximize the level of automation during the training of the network, some Fast.ai callbacks have been used to perform the early stopping of the training (with $patience = 10$, i.e. the training stops when the validation loss of the network does not improve for $10$ consecutive epochs) and to automatically save the best model of the current training round and then reload it for the validation and for the next round at higher resolution. 
Two consecutive training rounds were run, the first at 360 px resolution, the second at 540 px resolution, with a variable number of training epochs dependent on the early stopping callback of Fast.ai, and each composed of a \textit{freeze} and a \textit{unfreeze} step (training only the last output layer of the network or also all the convolutional layers). After each \textit{freeze/unfreeze} step is finished, the best model weights of the current step are re-loaded in memory, the original pre-training weights are restored and the training continues with the next step (i.e. next \textit{freeze/unfreeze} possibly at the next resolution).

Batch sizes were set to 8 ($360\times360$ px) and 4 ($540\times540$ px) for the \emph{freeze} and \emph{unfreeze} steps, respectively. The learning rates were set to $1e-03$ for the \emph{freeze} step and $slice(1e-07, 1e-06)$ for the \emph{unfreeze} step. The \emph{slice} notation is used to train the network with layer-specific learning rates \cite{layer-specific-learning-rates}. Finally, we train the model on the 3340 image/mask pairs in the training set.

\subsection{Semantic Segmentation of Crack and Pothole using Deep CNN and Learned Active Contours [\PUCP{}], by Miguel Chicchon and Ivan Sipiran}

\begin{figure*}[t]
\centering
\includegraphics[width=\linewidth]{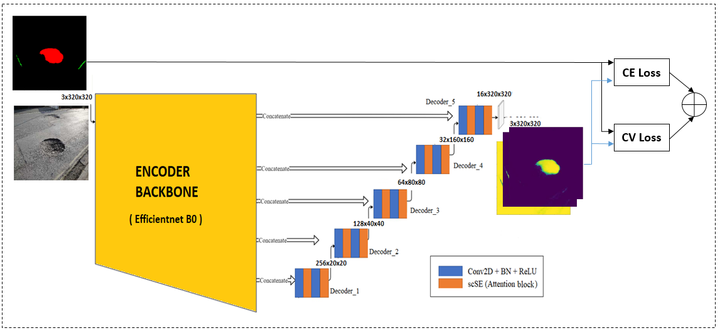}
\caption{Overview of the \PUCP{} method}
\label{fig:mich}
\end{figure*}

For this problem, the authors investigate the effect of a loss function $L$ based on active contour theory on deep neural network training. The implementation of the loss function corresponds to the representation through the Level Set method of the energy functional proposed by Chan-Vese~\cite{mich_1}. 

Experiments were performed combining the loss functions based on active contours~\cite{mich_2,mich_3,mich_4} and the cross-entropy loss, as follows:
\begin{equation}
    L=\alpha L_{CE} + \beta L_{CV}.
\label{eq1}\end{equation}

\begin{equation}
    L_{CE} = -\frac{1}{N}\sum_{n=1}^{N} \sum_{c=1}^{C} \sum_{p=1}^{P} T_{ncp} \ln\left ( Y_{ncp} \right ).
\label{eq2}\end{equation}
\begin{equation}
\begin{aligned}
    L_{CV}  & = \frac{1}{N}\sum_{n=1}^{N}\sum_{c=1}^{C} \Biggl( \sum_{p=1}^{P} \left | T_{ncp}-c_{ncp,1} \right |^{2}H_{\xi}\left(\phi_{ncp}\right)  + \quad\cdots \\
    &\quad \sum_{p=1}^{P} \left | T_{ncp}-c_{ncp,2} \right |^{2} \left ( 1 - H_{\xi}\left(\phi_{ncp}\right) \right )  \Biggr).
\end{aligned}
\label{eq3}\end{equation}
The parameters $\alpha$ and $\beta$ in Equation~\ref{eq1} are set to $0.1$ and $10$ respectively, as the best results are obtained with these values. Equation~\ref{eq2} represents the calculation of the cross-entropy as a function of the true pixels ($T_{ncp}$) and the predicted pixels ($Y_{ncp}$), where $n$ is the number of the image in the batch, $c$ is the class and $p$ is the number of pixels in the image. Finally, equation~\ref{eq3} represents the loss function based on the Chan-Vese functional~\cite{mich_1}, specifically the component of the internal and external region to the contour represented by the Level Set method.
\color{black}
The level set function $\phi$ is a shifted dense probability map that is estimated from $\xi_{ncp}=Y_{ncp}-0.5 \in [-0.5, 0.5]$, while $H_{\xi}$ is an approximated Heaviside function, defined by: 
\begin{equation}
H_{\xi}\left(\phi_{ncp}\right) = \frac{1}{2}\left [ 1+\frac{2}{\pi} \arctan\left ( \frac{\phi}{\xi} \right )\right ].
\end{equation}
The average intensity of binary ground truth map $T_{ncp}$ for contour inside and outside are:
\begin{align}
c_{ncp,1}\left ( \phi_{ncp}  \right ) & = \frac{\sum_{p=1}^{P}T_{ncp}H_{\xi}\left(\phi_{ncp}\right)}{\sum_{p=1}^{P}H_{\xi}\left(\phi_{ncp}\right)},\\ 
c_{ncp,2}\left ( \phi_{ncp}  \right ) & = 
\frac{\sum_{p=1}^{P}T_{ncp}\left ( 1 - H_{\xi}\left(\phi_{ncp}\right) \right )}{\sum_{p=1}^{P}\left ( 1 - H_{\xi}\left(\phi_{ncp}\right) \right )}.
\end{align}

State of the Art segmentation network architectures such as UNet, UNet++, MANet, LinkNet, FPN and DeepLabV3+ were experimented with pre-trained networks based on the Efficientnet architecture for the encoding stage. In all cases, the combined loss function allowed to improve the training results, selecting the 3 best models corresponding to the UNet++, MANet and UNet architectures. An overview of the method is shown in Figure~\ref{fig:mich}.


\subsection{From SegFormer to Masked Soft CPS [\HCMUS{}], by Minh-Khoi Pham, Thang-Long Nguyen-Ho, Hai-Dang Nguyen and Minh-Triet Tran}
\begin{figure*}[t]
    \centering
    \includegraphics[width=.85\linewidth]{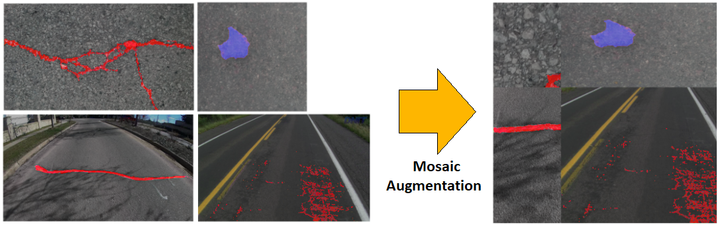}
    \caption{An example of the Mosaic Augmentation used in \HCMUS.}
    \label{fig:HCMUS_1}
\end{figure*}

The authors of this submission adapted well-known state-of-the-art models in segmentation, including UNet++ \cite{zhou2018unet++}, DeepLabV3+ \cite{chen2018encoder} and recent SegFormer \cite{xie2021segformer}, to the problem of the pothole detection. In particular, the authors used data augmentation to balance the situation where each image has only one class. Indeed, the main observation at the core of this proposal is that the data provided by the organizers only contain one of the two classes of damage (in most cases), however, real road scenarios usually have a large assortment of damage types in the same image. From that motivation, the authors augment the data by stitching the images together to simulate the cracks and the potholes appearing in the same scene. In particular, this is done via mosaic data augmentation to blend multiple images into a single one. This creates new simulated data that introduces a variety of possible situations where both cracks and potholes are present in the same scene. Figure~\ref{fig:HCMUS_1} shows an example of mosaic data augmentation.

Then, the authors ran different experiments with different augmentation and hyperparameters settings. However, all the three proposed setups share the same objective function. Initially, authors went for the Cross Entropy (CE) and Dice loss, since it is a common combination. This leads to poor recall metrics, so the authors guessed that the background pixels outnumbering pothole/crack pixels in most of the training samples and a number of inaccurate ground-truth masks are the reason behind this. Then, the authors focused on detecting as many road damages as possible, i.e.: they assumed that a higher recall would give more reasonable visual results than higher precision. This led to the adoption of a loss function which is a combination of Focal Tversky loss (FTL) \cite{abraham2019novel} and Cross Entropy with Online Hard Example Mining (OhemCE) loss (also known as Bootstrapping Cross Entropy Loss~\cite{OhemCE}). Details on these two loss functions can be found in the respective references, however, briefly: 
\begin{itemize}
\item Focal Tversky loss weights False Negative (FN) and False Positive (FP) by $\alpha$ and $\beta$ terms. Because authors wanted a high recall rate, they penalized the FN term more.
\item OhemCE only considers top-k highest loss pixels in the predicted masks. This helps the networks not to be overconfident in void pixels. We constrained the $k$ to be equal $H \times W \div 16$.
\end{itemize} 

Indeed, these two loss functions are nothing more than parametrized variants of the Dice/Cross Entropy loss respectively, adjustable to force the network to focus more on the recall score while maintaining fine accuracy, thus leading to better overall results. In particular, the FTL is: 
\begin{equation}
    FTL = (1 - Tl)^\gamma
\end{equation}
where Tl is:
\begin{equation}
    Tl = \frac{TP}{TP + \alpha FP + \beta FN}
\end{equation}

In the following, the three different setups for the different runs are described. Every solution builds over the knowledge acquired from the previous one, leading to the last run to be more developed.

\subsubsection{SegFormer}
\label{sec:segFormer}
For their first submission, the authors chose a Transformer model, as they gained its place among state-of-the-art recently. In particular, they used the SegFormer~\cite{xie2021segformer} model. The intention was both to check its performance in this scenario and also to assess the domain adaptation capabilities of the Transformer models family. However, the limitation in this architecture category is its slow convergence. In terms of implementation, the authors inherit a pre-trained model from the Huggingface library~\cite{huggingface}. 

\subsubsection{EfficientNet DeepLabV3+}
The authors trained the traditional DeepLabV3+~\cite{chen2018encoder} with some implementation changes. In particular, they reused the pre-trained EfficientNets \cite{tan2019efficientnet} on the ImageNet dataset as the backbone and train the whole architecture with fully-annotated labels. With this setup, the Dice score on the validation set increased from about 0.6 to 0.8 as verified on the test set by the track organizers. The Dice scores of this experiment are also good, once again demonstrating the efficiency of the DeepLabv3+ architecture in semantic segmentation problems.

\subsubsection{Masked Soft Cross Pseudo Supervision}
\label{sec:mscps}
The authors observed that while the setup described in Section~\ref{sec:segFormer} gave overall good metric scores on the validation set, it performed worse when it comes to out-of-distribution samples, such as frames from RGB-D videos. To fix this tendency, the authors strengthened the model with unsupervised data or rather data \emph{in-the-wild}. In particular, they utilized a non-annotated dataset (i.e.: only the RGB images without the masks and the frames of the RGB-D videos) for the unsupervised training branch, aiming at enhancing the capabilities of the model to predict out-of-distribution samples. 

\begin{figure}[t]
    \centering
    \begin{tabular}{c}
        Supervised branch\\
        \includegraphics[width=.9\linewidth]{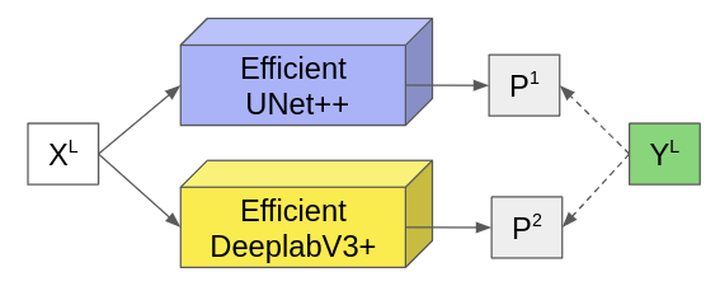}\\ \\ \\
        Unsupervised branch \\
        \includegraphics[width=.9\linewidth]{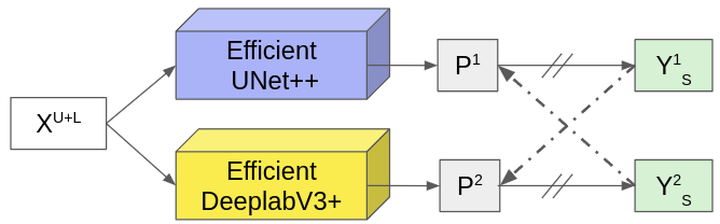}
    \end{tabular}
    \caption{Both branches of the setup of the \HCMUS{} method described in Section~\ref{sec:mscps}. $X^L$, $X^{U+L}$ indicates labelled inputs, unlabelled and labelled inputs respectively. $Y^{L}$ and $Y^{S}$ are segmentation masks (the ground-truth one and the soft pseudo one respectively) while $P$ means the probability maps defined by the networks. ($\rightarrow$) means forward, ($//$ on $\rightarrow$) means stop-gradient, ($-\rightarrow$) means loss supervision and ($-\cdot \rightarrow$) means masked loss supervision.}
    \label{fig:HCMUS_2}
\end{figure}

This setup is inspired by the recent semi-supervised method Cross Pseudo Supervision (CPS) \cite{chen2021-CPS}, with some critical improvements. Specifically, the authors softened the hard-coded pseudo labels with soft-max normalization and masked out the background channel (hence the name \emph{Masked Soft CPS}). Indeed, the original CPS method uses hard-coded pseudo labels and one-hot encoding to generate pseudo masks for dual training, which the authors thought would hurt performance on this dataset, as the type of model required to face this problem usually confidently predicts void pixels. Furthermore, annotated labels are not accurate perfectly, so if we use strict loss, which forces the model to learn the difference between foreground and background, it will lead to some confusion of prediction in contour positions. 
Moreover, the authors masked out the void pixel when training, so that these pixels are not counted in loss computation. CPS works by combining both the annotated and non-annotated data and training two neural networks simultaneously (DeepLabV3+ and Unet++ in our experiment). For the annotated samples, supervision loss is applied typically. For the non-annotated, the outputs from one model become the other's targets and are judged also by the supervision loss.
Figure~\ref{fig:HCMUS_2} illustrates this training pipeline.

In the inference stage, the authors employed the ensemble technique used in~\cite{filipiak2021n} by merging the two logits derived from both networks by getting the max probabilities out of them, then weighted the results by heuristic numbers. In particular, the logits of cracks are multiplied by 0.4, potholes by 0.35 and background by 0.25. These numbers mean that there is more focus on cracks damage since these are more difficult to detect.


\section{Evaluation environment}
\label{sec:results_discussion}
This section presents and discusses the performances of the proposed methods (plus the baseline). Quantitative and qualitative evaluations are presented in Section \ref{sec:eval}, then, the overall discussion of the performance for each method is provided in Section  \ref{sec:discussion}.

\subsection{Results}
\label{sec:eval}
To achieve fairness and parity in the evaluation procedure, we collected all 7 methods in a single Jupyter notebook. The hardware used is an Intel Core i9-9900K PC with 32 GB of RAM and an Nvidia GeForce RTX 2070 GPU with 8 GB of video RAM. This allows us to evaluate the performance of the different models using the same environment (i.e.: same code, data, metrics, initial conditions, etc.). The notebook is publicly available in the following formats:  \Bhref{http://deeplearning.ge.imati.cnr.it/genova-5G/notebooks/shrec-2022-evaluation.html}{html} and \Bhref{http://deeplearning.ge.imati.cnr.it/genova-5G/notebooks/shrec-2022-evaluation.ipynb}{ipynb}. 

In Table~\ref{tab:res_images_valid} and Table~\ref{tab:res_images_test} we summarize the performance of the 7 runs (one for each method) on the validation and test sets, respectively. There are no huge gaps between the scores of the different models, however, the runs "emph{\PUCP{}-Unet++}" and "\emph{\HCMUS{}-CPS-DLU-Net}" (in bold) stand out from the others. As can be seen in the tables, for many of the methods the score trend is similar in the results of both validation and test sets. This means that the training, validation and test sets are sufficiently homogeneous with each other and the models have learned to extract features correctly and to represent and model the underlying probability distributions.

A qualitative evaluation is performed on 8 video clips: 3 are top-down videos taken on foot, 1 is wide-view on foot and the others are wide-view shot from a car. We applied each DL model to every frame of the videos and overlayed the resulting mask onto the video for easier evaluation. In this evaluation of wide-view videos, we mostly ignore small false positives on trees and other elements. Indeed, with lane detection techniques, it is possible to limit the recognition to the road surface only. However, we consider this mislabelling as an issue if they happen consistently on a wide number of non-road elements. The videos are publicly available at the following hyperlinks, one for each run: \Bhref{http://deeplearning.ge.imati.cnr.it/genova-5G/video/shrec-22-evaluation/baseline/0.deeplabv3+-resnet101.html}{Baseline (DeepLabv3+)}, \Bhref{http://deeplearning.ge.imati.cnr.it/genova-5G/video/shrec-22-evaluation/submission-1-Team1/0.manet.html}{\PUCP{}-MAnet}, \Bhref{http://deeplearning.ge.imati.cnr.it/genova-5G/video/shrec-22-evaluation/submission-1-Team1/1.unet.html}{\PUCP{}-U-Net}, \Bhref{http://deeplearning.ge.imati.cnr.it/genova-5G/video/shrec-22-evaluation/submission-1-Team1/2.unetpp.html}{\PUCP{}-U-Net++}, \Bhref{http://deeplearning.ge.imati.cnr.it/genova-5G/video/shrec-22-evaluation/submission-2-Team2/0.segformer.html}{\HCMUS{}-Segformer}, \Bhref{http://deeplearning.ge.imati.cnr.it/genova-5G/video/shrec-22-evaluation/submission-2-Team2/1.deeplabv3plus.html}{\HCMUS{}-DeepLabv3+}, \Bhref{http://deeplearning.ge.imati.cnr.it/genova-5G/video/shrec-22-evaluation/submission-2-Team2/2.maskedsoftcps-dlunet.html}{\HCMUS{}-Masked SoftCPS DLU-Net}.
Overall, the performances of the runs vary: some methods perform better on some specific types of videos (e.g., methods very effective in top-down videos may become less so in wide-view videos). We detail the qualities of each method in the following section.

\begin{table}[t]
\footnotesize
    \centering
    \caption{Evaluation on the image validation set. Values range from 0 (red), to 1 (green). The higher the value is, the better the method performs. Most valuable runs are highlighted in bold.}
    \begin{tabular}{|c||*{5}{R|}}
        \hline
                                  &WPA      & DM    &mIoU   &cIoU   &pIoU\EndTableHeader\\ \hline \hline
        \baseline{} - DeepLabv3+  &0.682	&0.814	        &0.711	&0.606  &0.760		  \\ \hline
        \PUCP{}-MAnet             &0.774	&0.810	        &0.705	&0.719	&0.781	  \\ \hline
        \PUCP{}-Unet              &0.754	&0.804	        &0.698	&0.693	&0.776	  \\ \hline
        \textbf{\PUCP{}-Unet++}            &0.767	&0.800	        &0.694	&0.706	&0.801	  \\ \hline
        \HCMUS{}-SegFormer        &0.671	&0.637	        &0.523	&0.633	&0.624	  \\ \hline
        \HCMUS{}-DeepLabv3+       &0.727	&0.802	        &0.695	&0.642	&0.780  \\ \hline
        \textbf{\HCMUS{}-CPS-DLU-Net}      &0.840    &0.763          &0.647  &0.777  &0.864  \\ \hline
    \end{tabular}

    \label{tab:res_images_valid}
\end{table}

\begin{table}[t]
    \centering
    \footnotesize
    \caption{Evaluation on the image test set. Values range from 0 (red), to 1 (green). The higher the value is, the better the method performs. Most valuable runs are highlighted in bold.}
    \begin{tabular}{|c||*{5}{R|}}
        \hline
                                  &  WPA       & DM   &mIoU  &cIoU  &pIoU \EndTableHeader  \\ \hline \hline
        \baseline{} - DeepLabv3+  &  0.598     & 0.789        &0.676 &0.645 &0.584 \\ \hline
        \PUCP{}-MAnet             &  0.752     & 0.827        &0.725 &0.787 &0.728  \\ \hline
        \PUCP{}-Unet              &  0.741     & 0.824        &0.720 &0.776 &0.717  \\ \hline
        \textbf{\PUCP{}-Unet++}   &  0.758     & 0.832        &0.731 &0.780 &0.762  \\ \hline
        \HCMUS{}-SegFormer        &  0.802     & 0.747        &0.628 &0.763 &0.855  \\ \hline
        \HCMUS{}-DeepLabv3+       &  0.727     & 0.823        &0.719 &0.708 &0.818  \\ \hline
        \textbf{\HCMUS{}-CPS-DLU-Net}      &  0.833     & 0.789        &0.677 &0.843 &0.865  \\ \hline
    \end{tabular}

    \label{tab:res_images_test}
\end{table}

\begin{figure*}[t]
    \centering
    \footnotesize
    \begin{tabular}{|c|c|c|c|}
        \hline
        \baseline{} - DeepLabv3+	&\PUCP{}-MAnet	&\PUCP{}-Unet	&\PUCP{}-Unet++ \\	
        \includegraphics[width=.24\linewidth]{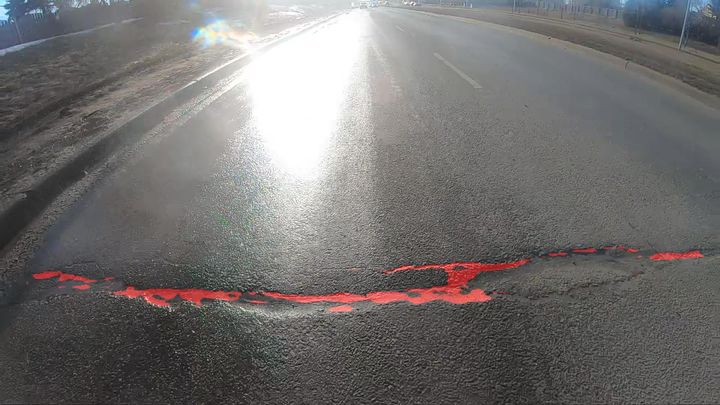}&
        \includegraphics[width=.24\linewidth]{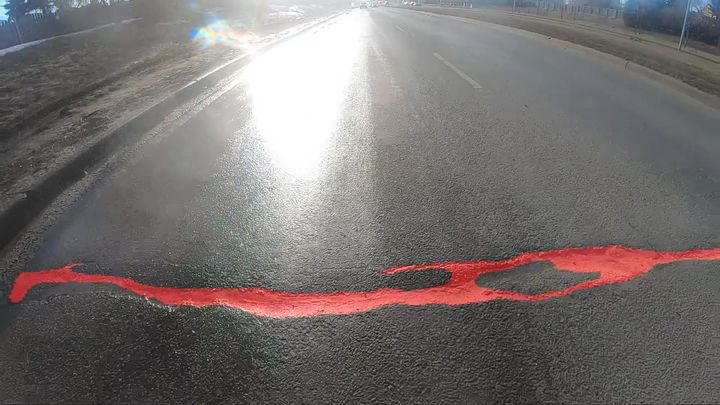}&
        \includegraphics[width=.24\linewidth]{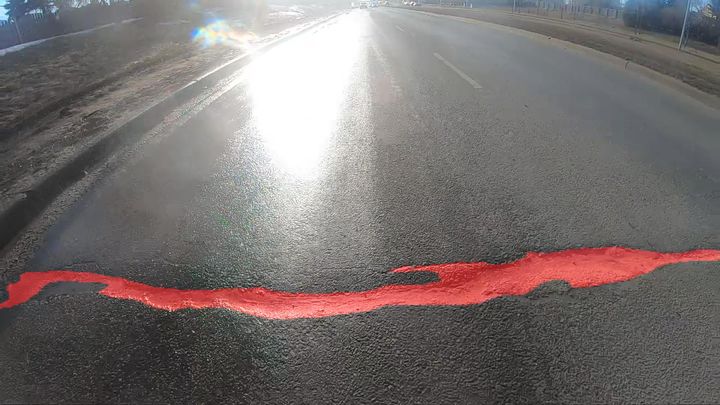}&
        \includegraphics[width=.24\linewidth]{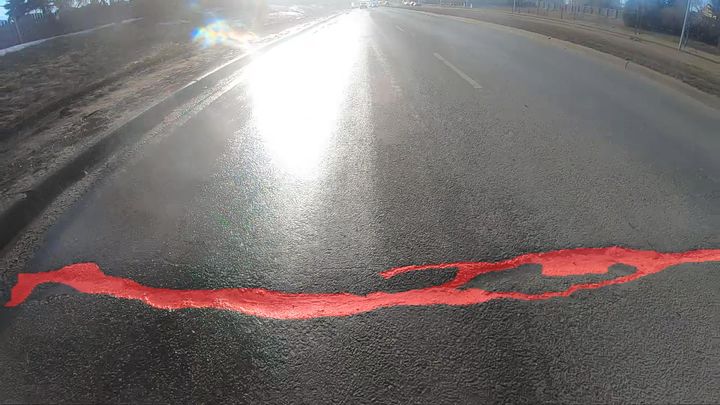}\\ \hline
        \multicolumn{4}{c}{ }\\ \hline
        \HCMUS{}-SegFormer	&\HCMUS{}-DeepLabv3+	&\HCMUS{}-CPS-DLU-Net & Ground-truth\\
        \includegraphics[width=.24\linewidth]{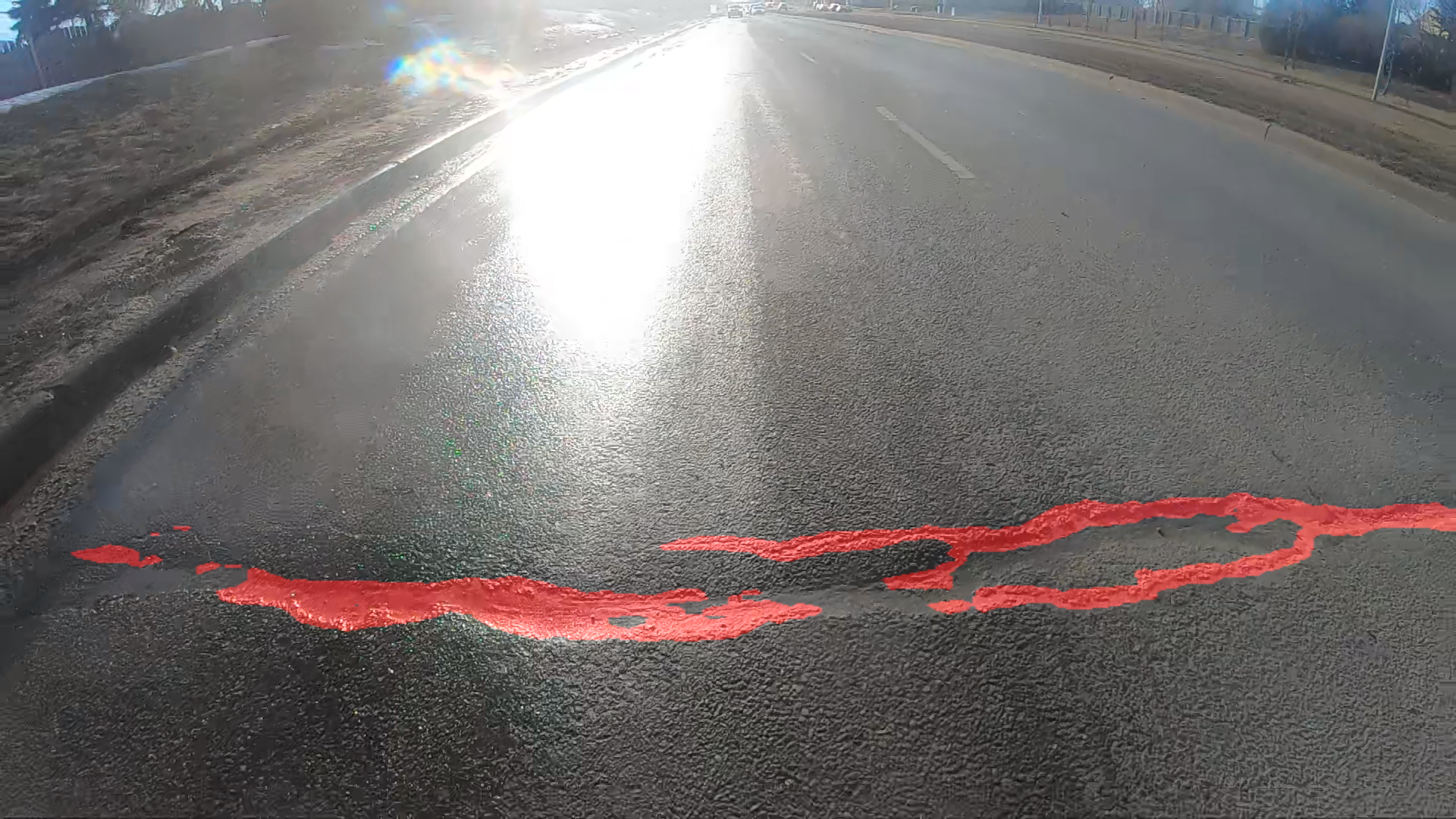}&
        \includegraphics[width=.24\linewidth]{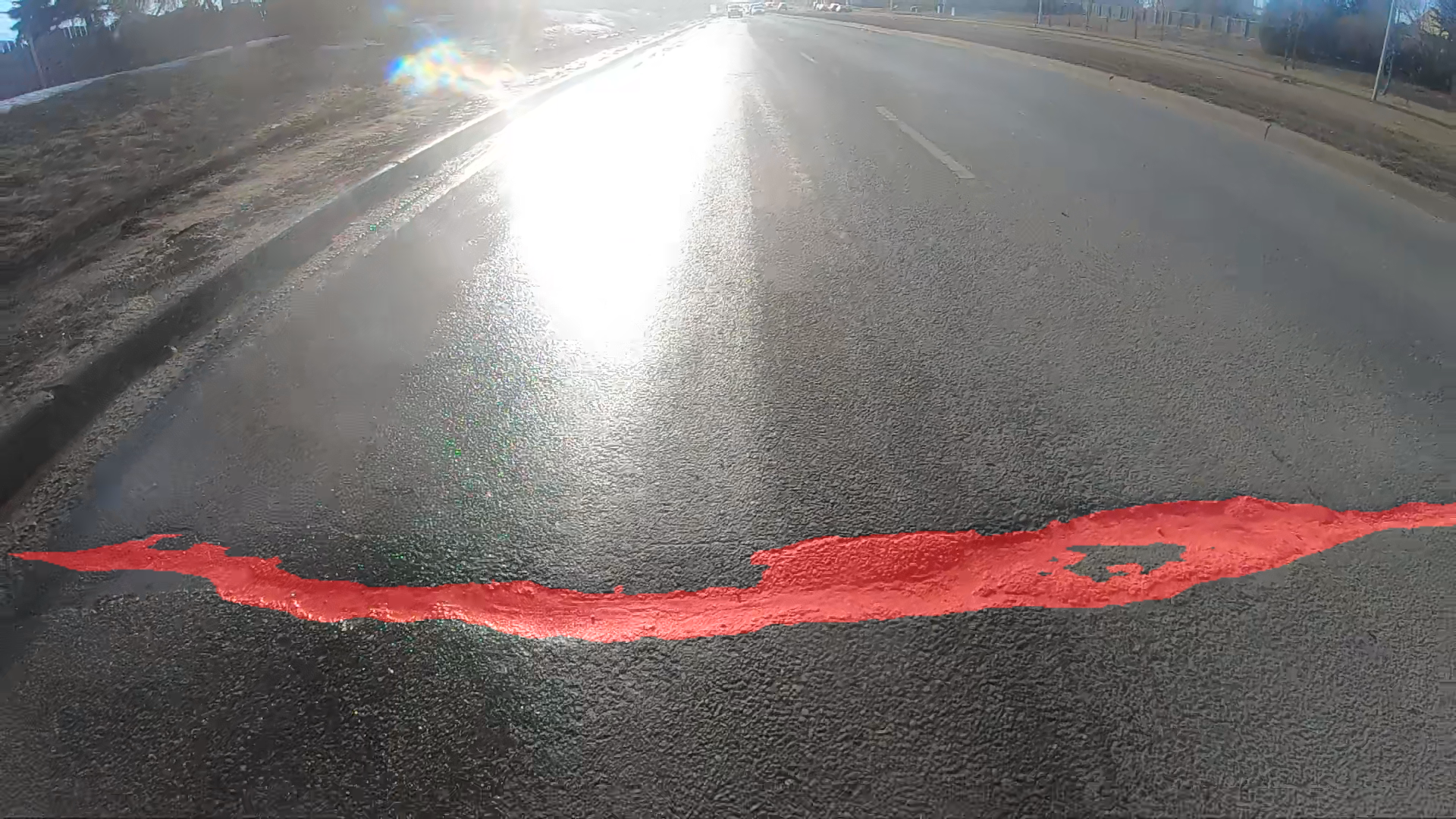}&
        \includegraphics[width=.24\linewidth]{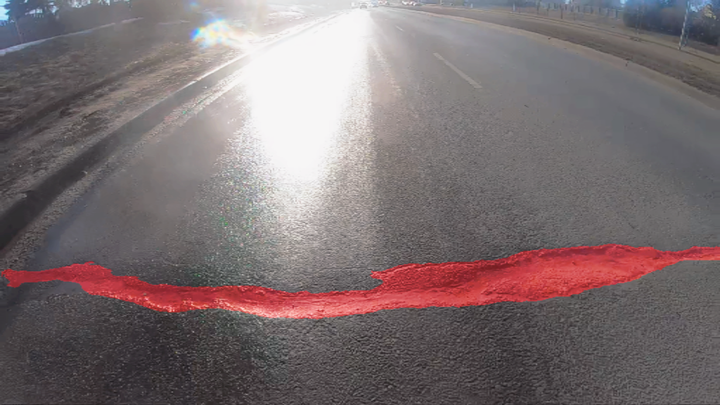} & 
         \includegraphics[width=.24\linewidth]{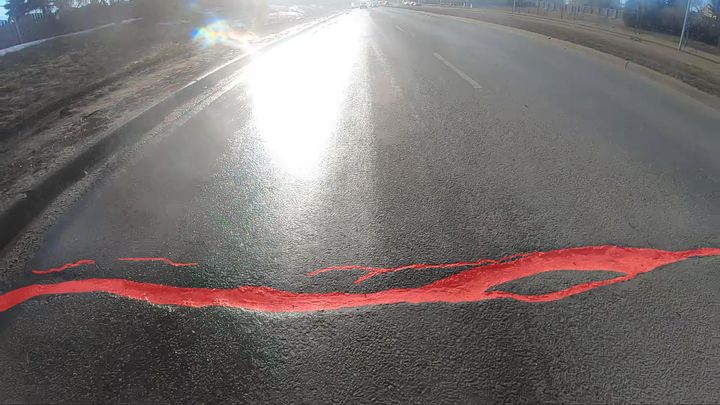}\\ \hline
    \end{tabular}
    \normalsize
    \caption{An example of the mask extracted by all the methods on a sample image representing a crack.}
    \label{fig:peculiar-masks-cracks}
\end{figure*}
\begin{figure*}[t]
    \centering
    \footnotesize
    \begin{tabular}{|c|c|c|c|}
        \hline
        \baseline{} - DeepLabv3+	&\PUCP{}-MAnet	&\PUCP{}-Unet	&\PUCP{}-Unet++ \\	
        \includegraphics[width=.24\linewidth]{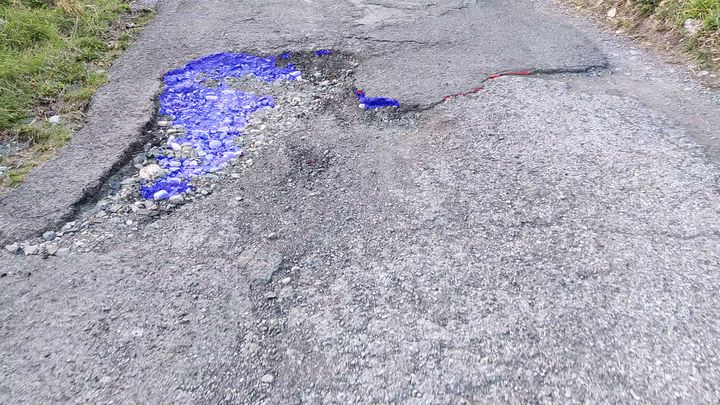}&
        \includegraphics[width=.24\linewidth]{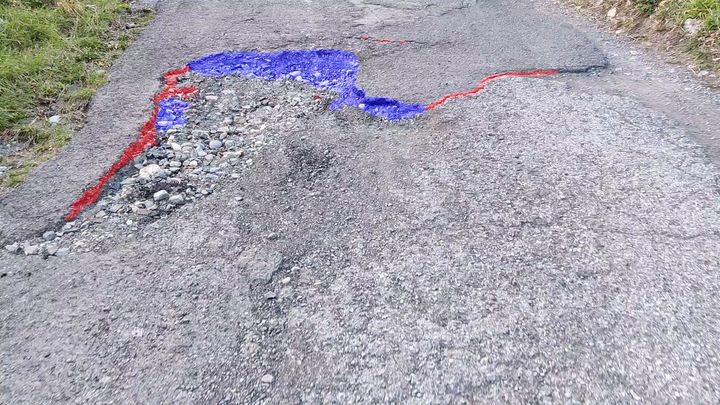}&
        \includegraphics[width=.24\linewidth]{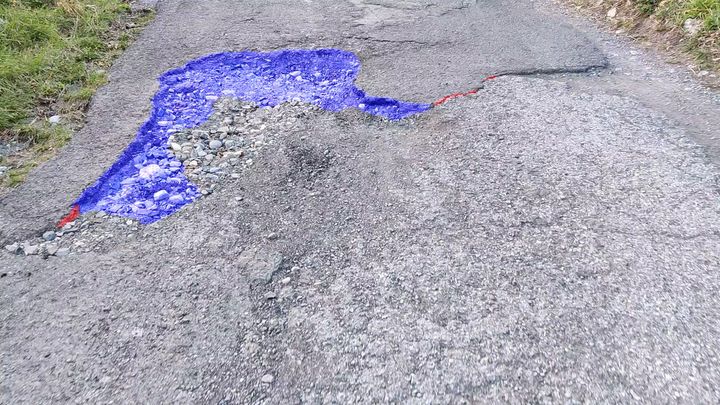}&
        \includegraphics[width=.24\linewidth]{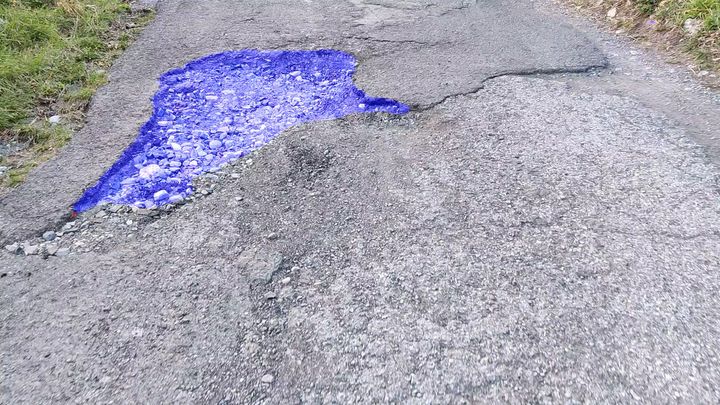}\\ \hline
        \multicolumn{4}{c}{ }\\  \hline
        \HCMUS{}-SegFormer	&\HCMUS{}-DeepLabv3+	&\HCMUS{}-CPS-DLU-Net & Ground-truth\\
        \includegraphics[width=.24\linewidth]{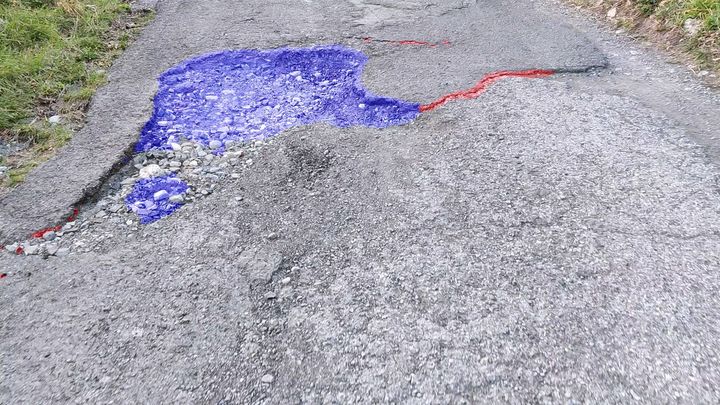}&
        \includegraphics[width=.24\linewidth]{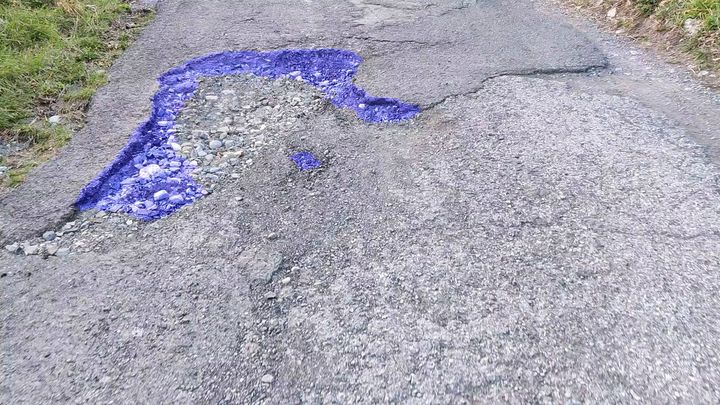}&
        \includegraphics[width=.24\linewidth]{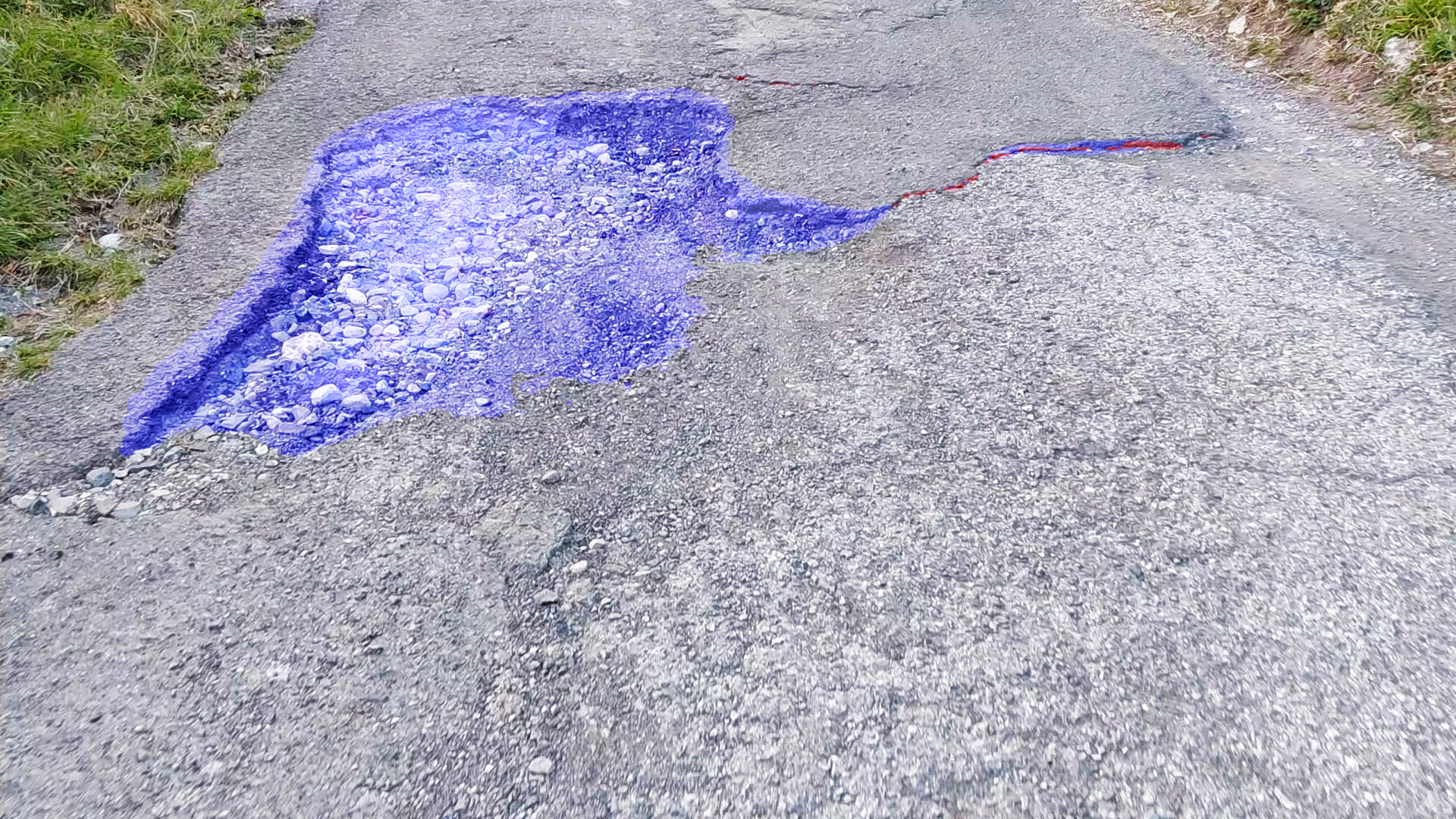} & 
        \includegraphics[width=.24\linewidth]{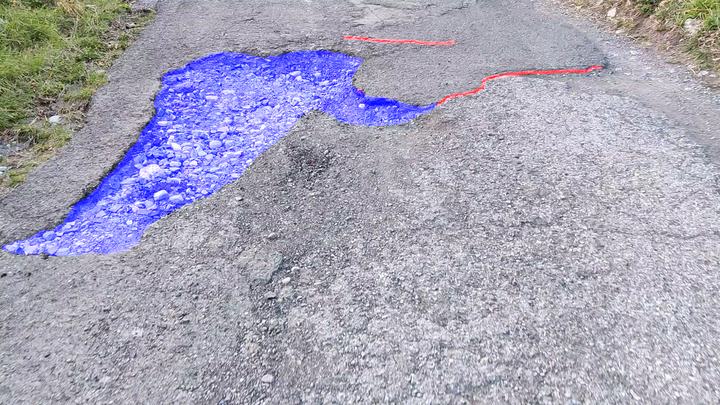}\\ \hline
    \end{tabular}
    \normalsize
    \caption{An example of the mask extracted by all the methods on a sample image representing a pothole and cracks.}
    \label{fig:peculiar-masks-potholes}
\end{figure*}

\subsection{Discussion}
\label{sec:discussion}
\begin{figure}[th]
    \centering
    \begin{tabular}{ccc}
        \includegraphics[width=.44\linewidth]{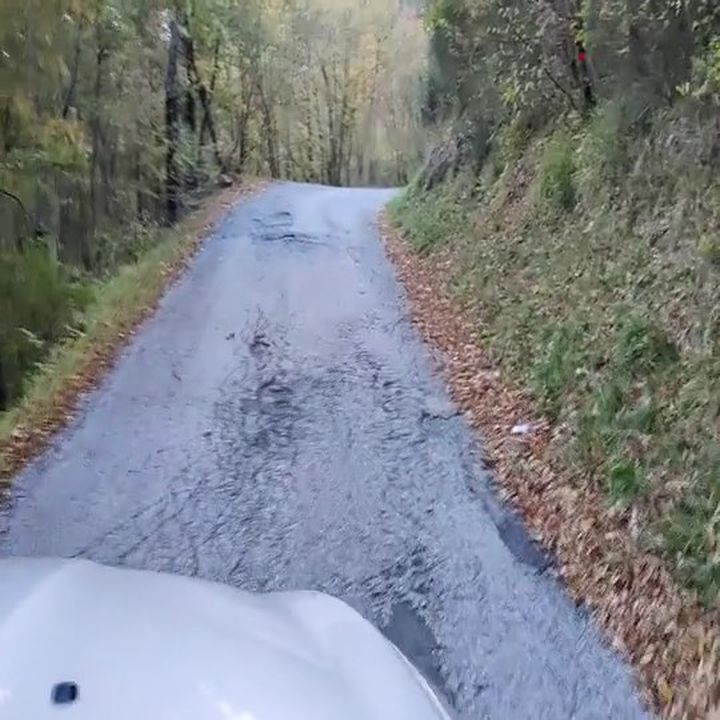}& $\qquad$ & 
        \includegraphics[width=.44\linewidth]{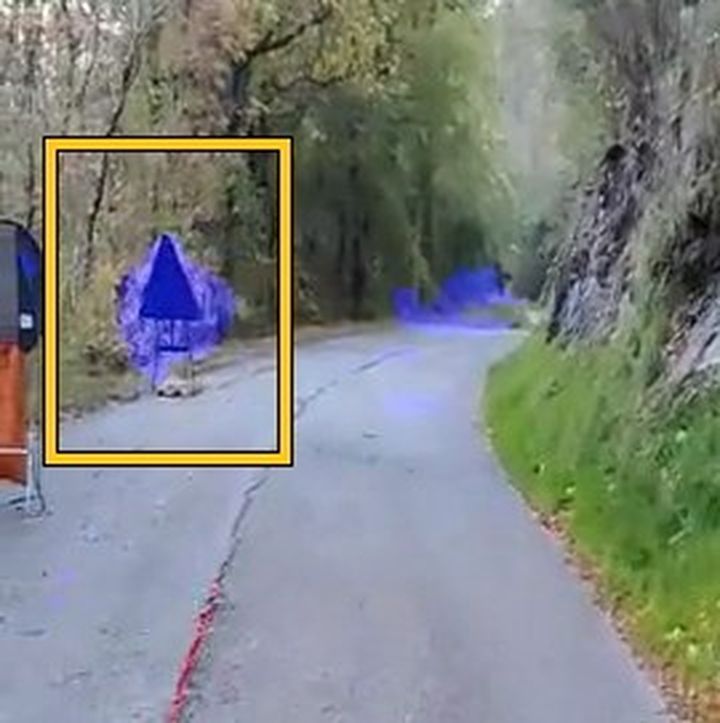}
    \end{tabular}
    \caption{Two frames extracted from the same video and used for qualitative evaluation. The masks of both frames were generated from \baseline{}. On the left, we show an example of \baseline{}'s lack of damage detection in wide-view video. On the right, we show how there is a strong correlation between the \baseline{}'s detection of a pothole and the presence of a dark blob of pixels. This last is not a complete frame but a zoom-in on one. For example, the traffic sign (yellow box) is recognized as a pothole.}
    \label{fig:baseline_vid_svw}
\end{figure}

The \baseline{} is able to detect most road damages but lacks in terms of the image segmentation quality. In other words, it scores high both true positives and false negatives. This is visible both in cracks and potholes (see Figure~\ref{fig:peculiar-masks-cracks} and ~\ref{fig:peculiar-masks-potholes}), in which the damage is spotted but the damaged surface is wider than the generated mask. This is especially evident with respect to the other methods mask on the same image. In the videos, especially in wide-view ones, this \textquotedblleft conservativeness\textquotedblright{} is sharpened and prevent \baseline{} from detecting most of road damages. Moreover, we observed false positives in correspondence of road signals and shadows. It could be argued that the detection of road damage is strongly related to the presence dark pixels. These last two issues are shown in Figure~\ref{fig:baseline_vid_svw}, in which we show two frames of a wide-view video: in one, \baseline{} spots no damages (left), in the other the back of a road signal is identified as a pothole (right).

\begin{figure}[th]
    \centering
    \begin{tabular}{ccc}
        \includegraphics[width=.44\linewidth]{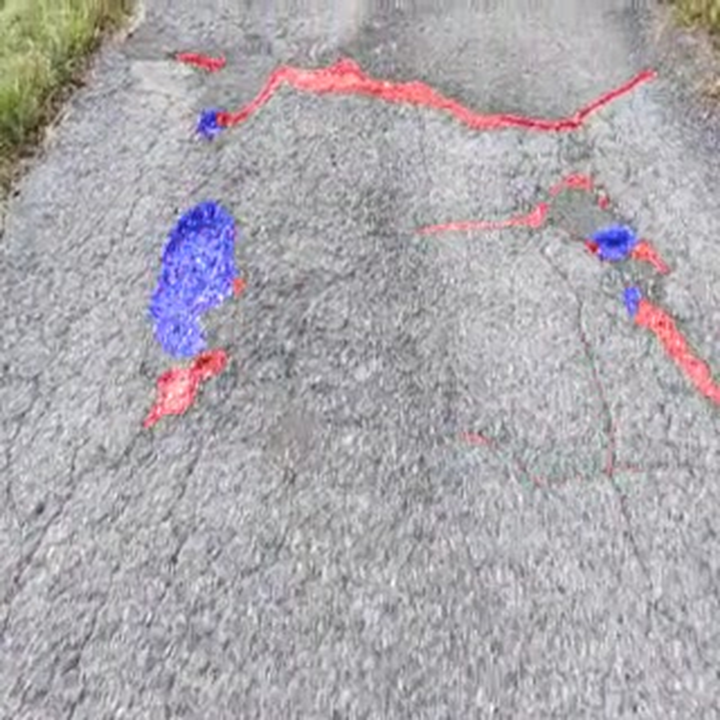}& $\qquad$ & 
        \includegraphics[width=.44\linewidth]{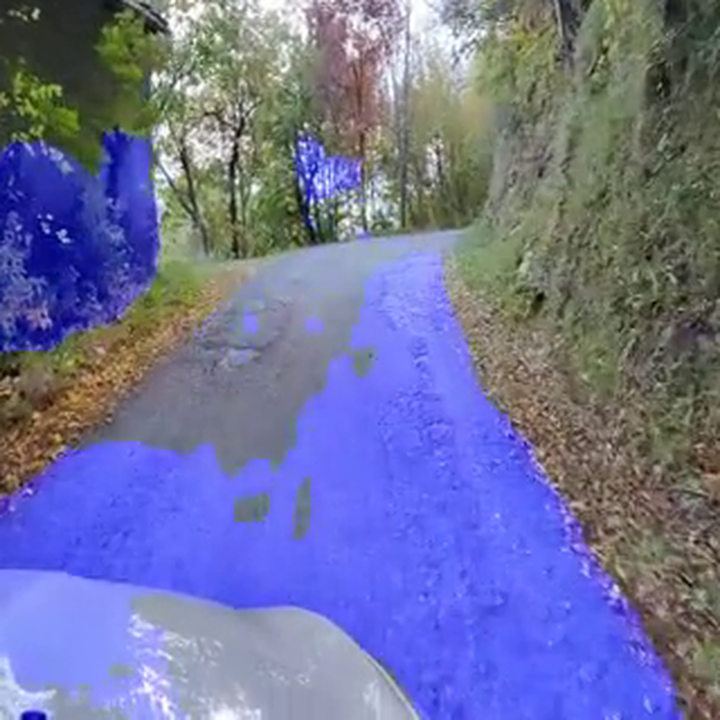}
    \end{tabular}
    \caption{Two frames extracted from two different videos and used for qualitative evaluation of \emph{\PUCP{}-MAnet} predictions. On the left, we show an example of its very good performance on top-down videos. On the right, the issues in predicting road damage on wide-view videos.}
    \label{fig:pucp_vid_svw}
\end{figure}

Regarding the \PUCP{} runs, the quantitative scores in Tables \ref{tab:res_images_test} and \ref{tab:res_images_valid} indicate that no run is significantly better than the others. This suggests that the value of the approach proposed by \PUCP{} is mainly in the loss function and data augmentation chosen rather than in the type of neural network architecture. Indeed, the Chan-Vese energy function~\cite{mich_1} takes into account global spatial information, whereas each prediction on pixels in a cross-entropy calculation is independent of the others. Furthermore, the representation of class predictions based on level set functions is more susceptible to global changes when small segmentation errors are present.
When analyzed on the videos, the \PUCP{} runs show consistent performances on the top-down videos, with great crack detection and segmentation accuracy. We evaluate \emph{\PUCP{}-MAnet} better than all the other runs of this contest for this type of videos. An example of this is shown in Figure~\ref{fig:pucp_vid_svw}(left). Nevertheless, wide-view videos contain a lot of false positives and mislabel, as shown in Figure~\ref{fig:pucp_vid_svw}(right). It is possible to conclude that using a loss function based on active contours improves the quality of shape or geometry segmentation, though it has little impact if the models fail to distinguish between classes well.

\begin{figure}[th]
    \centering
    \begin{tabular}{ccc}
        \multicolumn{3}{c}{\includegraphics[width=.99\linewidth]{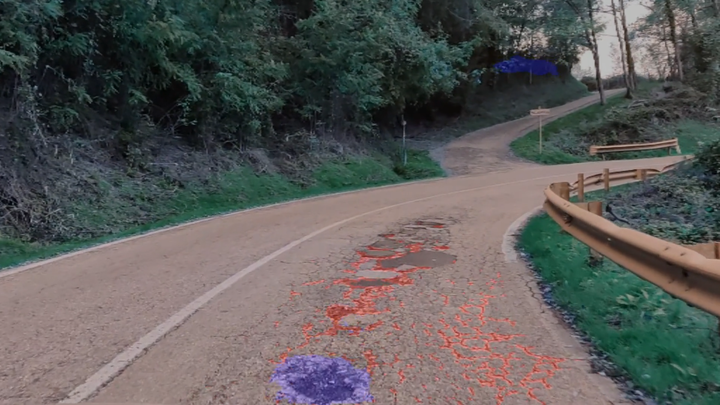}}\\
        \includegraphics[width=.33\linewidth]{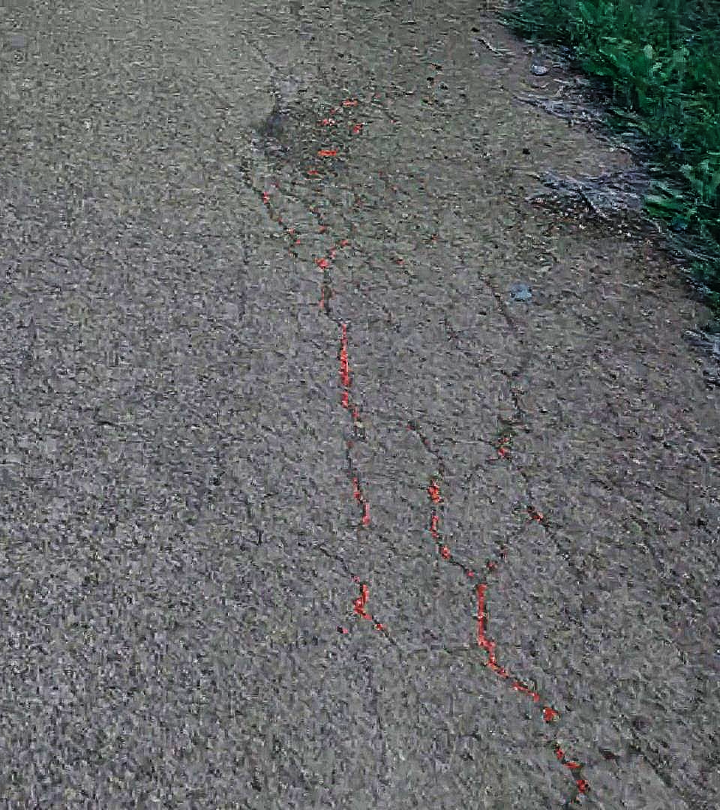} &
        \includegraphics[width=.33\linewidth]{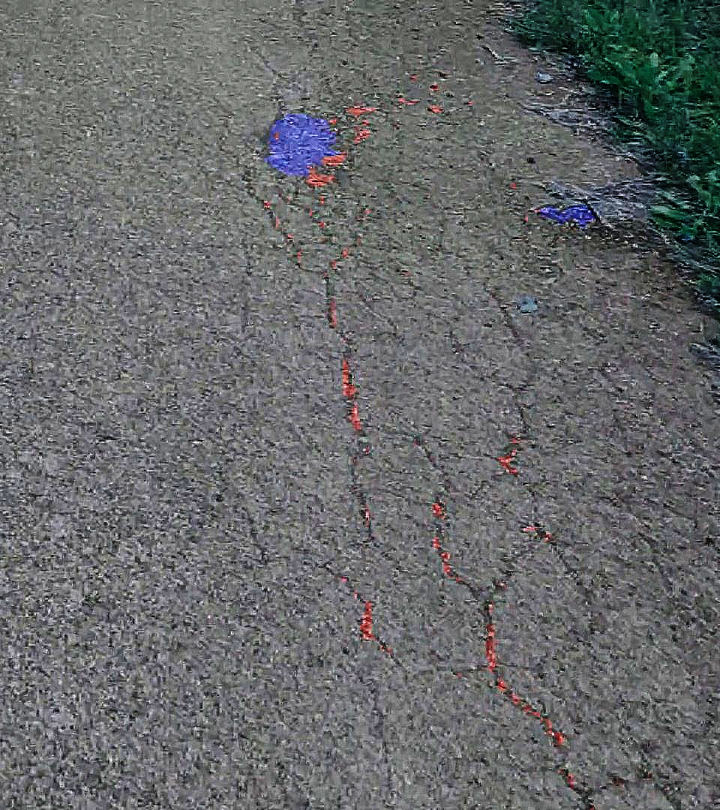} &
        \includegraphics[width=.33\linewidth]{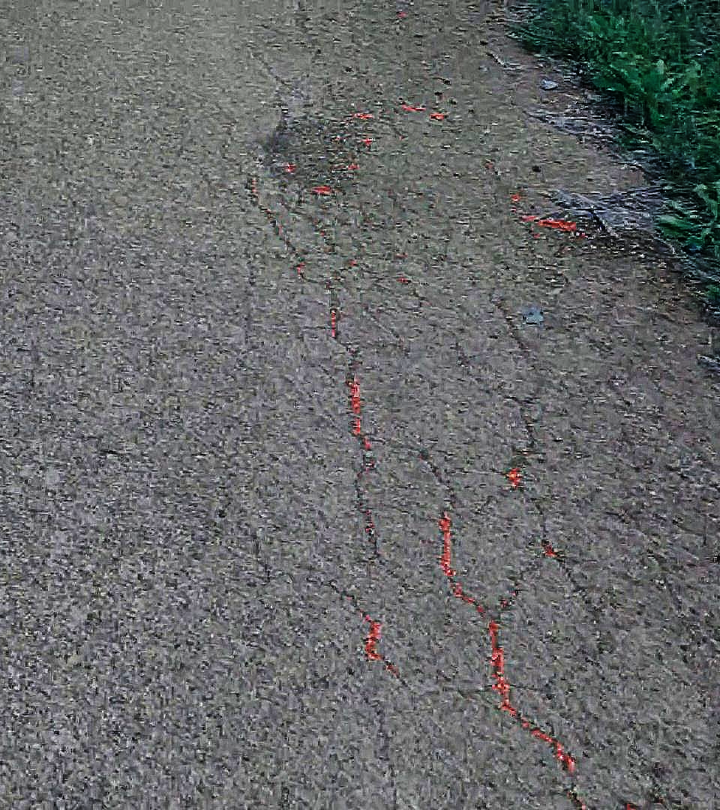}\\
    \end{tabular}
    \caption{Frames extracted from the videos used for the qualitative evaluation. Masks are generated by \emph{\HCMUS{}-CPS-DLU-Net+}. Top: an example of the good performance on wide-view videos. Bottom: 3 consecutive frames of a top-down video in which \emph{\HCMUS{}-CPS-DLU-Net+} segmentation varies significantly ("prediction flickering").}
    \label{fig:hcmus_vid_svw}
\end{figure}

\HCMUS{} outcome improves over the three runs, since they progressively refine the model (i.e.: \HCMUS{}-CPS-DLU-Net is on top of \HCMUS{}-DeepLabv3+ that is build on top of \HCMUS{}-SegFormer). Figure~\ref{fig:peculiar-masks-cracks} and ~\ref{fig:peculiar-masks-potholes} support this fact, as well as the results in Table~\ref{tab:res_images_valid} and ~\ref{tab:res_images_test}. It is interesting that the Dice Multi and mIoU evaluations drop significantly from \emph{\HCMUS{}-DeepLabv3+} to \emph{\HCMUS{}-CPS-DLU-Net} while the opposite happens for all the other evaluation measures. However, it is worth noticing that the CPS strongly focuses on the recall score therefore the model might be predicting too much of false positives. In that case, it reduces the overall score since the Dice and mIOU metrics take background pixels into consideration. In the videos, the potholes detection are great in both top-down and wide-view videos. Interestingly, distant potholes in wide-view videos are initially classified as cracks and then identified as potholes once the camera goes closer to them. Overall, \emph{\HCMUS{}-CPS-DLU-Net+} performs better on wide-view videos with respect to all the other runs of this benchmark (an example is shown in Figure~\ref{fig:hcmus_vid_svw} (top)) and obtains comparable results on top-down videos (despite being less efficient on cracks with respect to \emph{\PUCP{}-MAnet}). However, we notice less stability in the segmentation across consecutive video frames. An example is shown in Figure~\ref{fig:hcmus_vid_svw} (bottom) where three consecutive frames of one of the videos used for the qualitative evaluation are shown. Notice how both cracks and potholes are not constant from frame to frame, causing the typical "flickering" effect.
However, it is worth mentioning that this fact results as a downside with respect to the other methods mainly on cracks: indeed, this flickering effect occurs for all the methods when it comes to potholes.

Overall, \emph{\PUCP{}-Unet++} and \emph{\HCMUS{}-CPS-DLU-Net} stand out as the most valuable runs. In general but especially for the \baseline{} method, it is possible to notice that dark areas in the videos (like the back of a road sign or a decently dark shadow) are very likely to be mislabelled. Unfortunately, none of the participants exploited the information contained in the disparity channel of the RGB-D videos, that could help distinguish between shadow-like areas and actual change in the road surface. Only the method proposed in the run \emph{\HCMUS{}-CPS-DLU-Net} used data from RGB-D video clips, although it followed an unsupervised approach. The performance obtained with this run also exceeds those of the other runs submitted by the team.
\section{Conclusions and final remarks}
\label{sec:conclusions}

In this report we evaluated 7 methods (6 from the two participating teams, 1 provided by the organizers as a baseline) able to provide a solution to the ``SHREC 2022 track: pothole and crack detection in the road pavement using images and RGB-D data". All the methods submitted to this track are based on DL techniques.
In addition to supervised training on the training/validation sets of 3836 image/segmentation mask pairs provided by the organizers, the \HCMUS{} team chose an unsupervised approach to train one of their models using the RGB component of the provided RGB-D videos. However, none of the methods exploited the disparity map of the 797 RGB-D videos made available by the organizers. As per practice, the 504 image/mask pairs that made up the test set were not provided to the participants and were retained by the organizers for the final evaluation.

The methods submitted by the participants show very good results, both in quantitative and qualitative terms on the test videos (also not disclosed to the participants), despite performing differently based on the kind of test image/video. The final assessment of the organizers is that the two methods \emph{\PUCP{}-Unet++} and \emph{\HCMUS{}-CPS-DLU-Net} stand out as the most valuable runs.

In the future, it could be interesting to explore the possibility of having a dataset entirely built on RGB-D data and to exploit the whole data (i.e.: three color channels and the disparity map) to further help neural network models to better recognize road damage. Indeed, since many errors were found in correspondence of dark spots in the RGB images, the additional dimension can help the models to focus more on actual road surface disruption instead of color changes.
In parallel, the depth dimension could also help in the pretraining phase: using the disparity images as a label (possibly after a slight denoise/smoothing) should force the network to learn as many features as possible within the dataset, providing a possibly better basis for fine-tuning than a model pretrained on ImageNet.

\section*{Acknowledgements}
This work has been partially developed in the MISE Funded Project 5G Genova and in the CNR research activity DIT.AD007.041.002.

The work of Ivan Sipiran has been funded by Fondo Nacional de Desarrollo Científico, Tecnológico y de Innovación Tecnológica (FONDECYT) - SENCICO (Grant N° 129-2018-FONDECYT). 
The work of Miguel Chicchon has been funded by National Program for Innovation in Fisheries and Aquaculture (PNIPA) (PNIPA-ACU-SIA-PP-000588) and the Institute of Scientific Research (IDIC) of the University of Lima, Perú.

The organisers would like to thank Michela Spagnuolo for encouraging this activity and for her advice during the contest design phase.

\bibliographystyle{abbrv}

\end{document}